\documentclass[]{interact}
\usepackage{lineno,hyperref}
\usepackage{apacite}
\usepackage{epsfig} 
\usepackage[round]{natbib}
\usepackage{multirow}
\usepackage{graphicx}
\usepackage{array}
\usepackage[caption=false]{subfig}
\usepackage{amsmath}
\usepackage{float}
\usepackage{placeins}
\usepackage[justification=centering]{caption}
\usepackage{amsmath}
\usepackage{breqn}
\usepackage{epstopdf}
\usepackage[caption=false]{subfig}
\bibpunct[, ]{(}{)}{;}{a}{,}{,}

\theoremstyle{plain}

\theoremstyle{definition}

\theoremstyle{remark}

\begin{document}

\articletype{ARTICLE}

\title{Machine learning based modelling and optimization in hard turning of AISI D6 steel with newly developed AlTiSiN coated carbide tool}

\author{
\name{Anshuman Das\textsuperscript{a}, Sudhansu Ranjan Das\textsuperscript{b}, Jyoti Prakash Panda\textsuperscript{a}\thanks{Corresponding Author Email: jppanda@dituniversity.edu.in}, Abhijit Dey\textsuperscript{c}, Kishor Kumar Gajrani\textsuperscript{d}, Nalin Somani\textsuperscript{a}, Nitin Kumar Gupta\textsuperscript{a}}
\affil{\textsuperscript{a}Department of Mechanical Engineering, DIT University, Dehradun,
 Uttarakhand, India;\\
\textsuperscript{b}Department of Production Engineering, VSSUT, Burla, Odisha, India;\\
\textsuperscript{c}Department of Mechanical Engineering, NIT, Srinagar, Jammu and Kashmir, India;\\
\textsuperscript{d}Department of Mechanical Engineering, IIITDM, Kancheepuram, India}
}

\maketitle

\begin{abstract}
In recent times Mechanical and Production industries are facing increasing challenges related to the shift toward sustainable manufacturing. In this article machining was performed in dry cutting condition with a newly developed coated insert called AlTiSiN coated carbides coated through scalable pulsed power plasma technique in dry cutting condition and a dataset was generated for different machining parameters and output responses. The machining parameters are speed, feed, depth of cut and the output responses are surface roughness, cutting force, crater wear length, crater wear width and flank wear. The data collected from the machining operation is used for the development machine learning (ML) based surrogate models to test, evaluate and optimize various input machining parameters. Different ML approaches such as polynomial regression (PR), random forest (RF) regression, gradient boosted (GB) trees and adaptive boosting (AB) based regression are used to model different output responses in the hard machining of AISI D6 steel. The surrogate models for different output responses are used to prepare a complex objective function for the germinal center algorithm based optimization of the machining parameters of the hard turning operation.
\end{abstract}


\section{Introduction}
Numerous manufacturing industries worldwide are working on smart manufacturing techniques for improving productivity. Various new techniques are being tested for enhancement of productivity according to the industrial revolution, Industry 4.0. Some of these techniques are the Internet of Things (IoT), Machine Learning, Cyber-Physical system, big data, machines with computer controls and sensors and additive manufacturing \citep{zhang2020predicting,akhil2020image,mishra2020design}. By the applications of the above mentioned techniques, the production level has shown increments. Among all the mentioned techniques, IOT and big data are the most popular techniques \cite{surleraux2020machine}. The industries  4.0 are working on the information systems based on the sensors attached with the machines \citep{bricher2020supervised}. The manufacturing units are established and synchronized with various information systems through IOT and utilized in different production management purposes \citep{o2020determination}. Cyber physical systems allow the data acquisition in the real world which supports smart manufacturing. Nowadays,  the research in Machine Learning is flourishing because of large amounts of data that have been accumulated by various industries such as production, health, chemical, manufacturing and information technology (IT). Machine learning plays a vital role for enhancement in demand for artificial intelligence(AI) \citep{nagargoje2021performance}. It is a sub branch of AI, which allows the equipment to learn, improve and perform the specific task without disturbing the program. There are five steps involved in machine learning to solve a problem: problem definition, data collection, modelling, evaluation and result analysis\cite{pandamodeling}.

There are three broad categories of machine learning algorithms: supervised learning, unsupervised learning, and reinforcement learning. In supervised learning, data are trained for mapping between input and output. Knowing the ideal output for every sample in the input data acts as a guide for the learning procedure, thus providing supervision. In unsupervised learning, the ideal output is unknown and the model can only learn patterns present in the input data. In reinforcement learning, a predefined signal will be given, depending upon this, the machine will quantify the objectives. These broad ML approaches can be used for various tasks, such as clustering, dimensionality reduction, surrogate modeling for classification or regression, etc. ML based data-driven surrogate models can act as Digital Twins (DT) for real-life systems and these can be used for system exploration and optimization with a low computational expense.

At present, industries are conducting smart machining rather than conventional machining. Because errors are more severe in conventional machining, production is not effective and less efficient. In smart manufacturing, efficient production can be achieved with a stipulated processing time without hampering the quality of the product. In smart manufacturing, parameters can be optimized in real time through different subsystems and machines like controllers, sensors, systems, and machine tools. Numerous researchers have applied data-driven analysis for machining and production-based problems. The objective of these studies spans from using data to develop a deeper understanding of the system, to developing ML based models to act as surrogates for the system. \cite{kothuru2018application} determined the wear of the cutting tool and its failure using sound signals with the application of Support vector machine (SVM). The  tool wear was predicted accurately through the ML algorithm for different machining conditions. \cite{carrino2020machining}{} predicted the machining quality using ML techniques. The convolutional neural network (CNN) was used to predict and monitor the machined surface quality and the results showed that by combining both techniques resulted in 94\% accuracy of prediction. \cite{madhusudana2016fault} studied the fault diagnosis of face milling using the ML approach. For the identification of significant parameters, a decision tree algorithm was used. Also, the Naive Bayes model was used for the prediction of fault. It was observed that using the model, the results are predicted with 96.9\% accuracy. The ML models developed can be utilized for online monitoring of tool conditions and diagnosis of the fault of the tool used in milling operation. \cite{wang2019artificial} predicted the responses of wire electrical discharge machining (WEDM) through voltage signal using AI technique. From the analysis, it was found that the unsupervised AI techniques were predicted. \cite{krishnakumar2018machine} predicted the tool condition in a high-speed milling process using the ML technique. Four types of ML algorithms are employed, those are artificial neural network (ANN), decision trees, naive Bayes, and SVM. The SVM was found to be the most efficient method to predict the tool condition compared to other techniques. \cite{zhang2020feature} predicted the output responses in laser machining using deep multi-task learning. In this experiment, it was observed that Alex Net with multi-task learning was found to be better than deeper models. \cite{wang2020physics} detected the geometrical defects in WEDM using a physics-guided ANN model. The inconsistencies in the model were eliminated using a physics guided loss function, from the analysis it was observed that the predicted results matched well with the experimental results with 80\% accuracy. \cite{cho2005tool} detected the tool breakage in a milling operation using a support vector-based machine learning approach. From the experimental outcomes, it was confirmed that with the said technique, the machine downtime and cost of production can be effectively reduced compared to other techniques. \cite{wang2019heterogeneous} predicted the tool condition through the data-driven hybrid ML approach during milling of H13 steel and Inconel 718. From the results, it was found both the tool wear and surface roughness were predicted effectively with an accuracy of 85\% and 90\%. \cite{liu2020calibration} predicted the specific cutting energy with a hybrid approach of integrated ML technique. Both data-driven ML and process mechanics have been hybridized to predict the response. The results are well predicted and the authors claimed that the above-said model can be implemented in other cutting processes. \cite{chiu2017prediction} predicted the machining accuracy and surface quality using a data-driven approach by (Adaptive Neuro-Fuzzy Interface System) ANFIS model. From the simulation and experimental results, it was found that results can be effectively predicted through the data-driven model for better quality and productivity. McLeay et al.\cite{mcleay2020novel} detected the fault during the machining process using an unsupervised ML approach. Through the PCA plot, the fault conditions are observed. \cite{parwal2021machine}  used the machine learning-based approach to predict the tool wear during the machining operation. Various models are employed using logistic regression. The model was found to be good and results are interpreted effectively. And the results are suitable for industrial application. 
\cite{gouarir2018process} predicted the tool wear through the machine learning techniques. From the results, it was confirmed that the tool wear was predicted effectively with machine learning techniques and the accuracy was 90\%.
\cite{wu2017comparative} compared the tool wear prediction results through different ML techniques. Three types of algorithms are used like feed-forward back propagation(FFBP) based ANNs, random forest (RF) and support vector regression  (SVR). Better results are predicted using the RF algorithm.
\cite{fang2016neural} used a new artificial intelligence approach for the prediction of machined surface roughness in metal machining. ANN model was implemented. From the ANN model, it was observed that the machined surface roughness can be effectively predicted through the ANN model.
\cite{ulas2020surface} predicted the surface roughness of the machined surface of Aluminum alloy during WEDM with different machine learning algorithms. Four types of machine learning algorithms are applied such as ELM, W-ELM, SVR, and Q-SVR. Among four algorithms, weighted extreme learning machines (WELM) performed better.
\cite{bustillo2021machine} predicted the flatness deviation considering the wear of the face mill cutter teeth using a machine learning-based algorithm. Four different types of machine learning techniques were proposed, out of which, Random forest ensembles combined with the Synthetic Minority oversampling technique performed better. \cite{patange2021machine} predicted the health of a vibration-based multi-point tool insert on a vertical machine center using a machine learning approach. A tree-based algorithm was proposed by the authors. Various tree-based algorithms are used, out of which, the J48 decision tree-based algorithm was found to be the effective one.
\cite{oleaga2018machine} predicted the chatter in heavy-duty milling operation using machine learning techniques. Different machine learning models are used out of which, random forest performed better. And the authors concluded that with the use of machine learning techniques, dynamic machining performance can be enhanced in real working conditions. 
\cite{peng2019hybrid} predicted the cutting forces with consideration of the tool wear using machine learning methods. Two models are used one is conventional linear regression and the other is a hybrid model with machine learning. From the results, it was observed that the hybrid model with machine learning performed better.
\cite{mohanraj2021development} developed a tool condition monitoring system at the end milling process using wavelet features and statistical features based machine learning approach. Four different types of machine learning techniques are used to predict the flank wear like SVM,, K-nearest neighbor  (KNN), knowledge based (KB), decision tree, and MLP, and among all, SVM and DT performed better. \cite{sanchez2018unexpected} predicted the thickness change of machined parts during WEDM machining using machine learning techniques. Various models are used to predict the results. Among various models, a first convolutional layer with two gated recurrent units outperformed the other models. They also concluded that with the large data set, better results can be predicted. \cite{shen2020predicting} predicted the tool wear size in multi-cutting conditions using advanced machine learning techniques. Different models are used to predict the tool wear and the results are compared with the experimental results. It was observed that there was a good agreement between predicted and experimental results. Moreover, the authors studied that machine learning techniques can be used to predict the other machining responses. \cite{shastri2021optimization} optimized the process parameters in the machining of Titanium alloy in an MQL environment using an artificial intelligence-based algorithm. From the experimental results, it was observed that better results in terms of cutting force, tool wear, tool-chip contact length, and surface roughness were obtained using artificial intelligence-based optimization compared to PSO and the experimental approach. \cite{cheng2020intelligent} predicted the tool wear based on machine learning during the turning of high-strength steel. Two prediction models are used one is grid search algorithm-based support vector and the other is genetic algorithm-based support vector regression. From the results, it was found that with the tool wear, the stages of tool wear in a complete cycle can be easily predicted through machine learning techniques. Moreover, they suggested that through the machine learning method, tool wear monitoring can be made online. \cite{elsheikh2021fine} predicted the residual stress in turning of Inconel-718 using fine tuned AI model using pigeon optimizer. Two models were incorporated like hybrid ML model and traditional ANN  model. The traditional ANN was incorporated with three optimization techniques, those are bio inspired optimization, pigeon and Particle swarm optimization. The prediction accuracy of various models was examined through statistical measures. The traditional ANN was out performed by both ANN-PSO and ANN-PAO. \cite{jurkovic2018comparison} compared different machine learning methods for cutting parameter prediction in a high-speed turning process. Three types of ML techniques are compared i.e. SVR, Polynomial regression, and ANN. Cutting force, roughness, and machining time were the output responses. Better results in terms of cutting force and roughness were achieved with polynomial regression. Better machining time was predicted through ANN. The author also observed that for better prediction of results, more data set are required. \cite{cica2020predictive} predicted various machining responses under different cooling conditions in sustainable machining of 1045 steel using machine learning techniques. From the results, they revealed that an acceptable range of results can be predicted by using machine learning techniques without conducting actual experiments. So both time and money can be saved by using machine learning techniques. They also revealed that if a wide range of machining conditions will be adopted, better results can be predicted through this method. \cite{khoshaim2021optimized} predicted both mechanical and micro-structural properties of friction stir processed aluminium reinforced material using Grey Wolf optimizer. The input parameters are rotational speed, linear processing speed and number of passes, while the outputs were grain size, aspect ratio, micro-hardness and ultimate tensile strength. The prediction accuracy of developed hybrid model was found to be more accurate compared to standalone model.

\cite{zhao2020specific} predicted the specific energy consumption of machine tool based on tool wear in a dry milling operation. Three power characteristics are chosen like, standby power, cutting material power and no load power. Three responses are chosen spindle speed, MRR and tool wear. Both the process parameters and tool conditions are optimized for the reduction in energy consumption of machine tool. \cite{ravsovic2021recommended} recommended the layer thickness in a powder based additive manufacturing using multi attribute decision support. Many attributes are responsible for the product quality. But author has chosen two criteria one is strength and other is surface roughness. The most suitable layer thickness was proposed by three techniques called weighted features, DFAM knowledge and multiple attributes.
\cite{kahya2021precision} performed the precision and energy efficient machining of Ti6Al4V alloy on a turn mill machine tool. Three operations are conducted like face turning, rough flank milling and finish milling. Three responses are analyzed known as specific cutting energy, roughness and material removal rate. Different angle of inclinations (both lead and tilt) was analyzed on the response. The optimization technique employed was Particle swarm optimization. From the experimental outcomes, it was found that both lead and tilt angle influenced surface roughness effectively.
\cite{guo2021analysis} combined two methods DPCA and IMODE to analyze the effect o9f input variables on the gear quality in a hobbing process. The simulation of gear hobbing data was done and it was compared with NSGA, NSGAII and MODE. From the outcomes it was found that the proposed algorithm was very effective regarding diversity and optimization ability.
\cite{ruiz2020machine} predicted the tensile strength of the steel rods by machine learning algorithms those are manufactured in an electric arc furnace. Various ML algorithms are proposed like multiple linear regression, K-Nearest Neighbors, Classification and Regression Tree, Random forest, Adaboost gradient boost algorithms and ANN. Better results are predicted through fine-tuned random forest. And chemical variables are observed to be the most important variable affecting the material strength.
\cite{tian2020integrated} optimized the cutting parameters and processing sequence to minimize carbon emissions and process time in a CNC machine using an integrated multi objective optimization technique. The modelling of process parameters is done using NAGS-II technique. From the results, it was found using integrated multi objective optimization technique, both processing time and carbon emission can be minimized effectively.

\cite{saidi2019modeling} used response surface methodology for modeling the output machining responses and optimized the input parameters using a desirability function approach.\cite{jumare2019predictive} used a similar approach as of \cite{saidi2019modeling} for modeling and optimization of the machining process parameters in ultra-high precision machining of optical lenses. Caggiano et al.\cite{caggiano2018machine} optimized the tool life exploitation during CFRP composite drilling in aeronautical assembly using ML techniques. From the experimental outcomes, it was found that both fractal and statistical analysis can predict better results. \cite{shukla2019application} applied ML techniques for optimization in WEDM of Haste alloy C-276. From the experimental outcomes, it was observed that both surface roughness and kerf width were influenced by pulse on time, pulse off time, and peak current. And the gradient descent method was successfully implemented for the prediction of optimum results. \cite{nain2018evaluation} analyzed the cutting speed, wire wear ratio, and dimensional deviation of WEDM machining of superalloy Udimet L605 using both support vector and grey relational analysis. From the experimental results, it was observed that pulse on time was the significant variable affecting the responses. The highest percentage of Copper (8.66\%) was obtained during machining at the highest cutting speed, whereas with low cutting speed 7.05\% of Copper was observed. 

\begin{figure}
\centering
\includegraphics[width=0.95\textwidth]{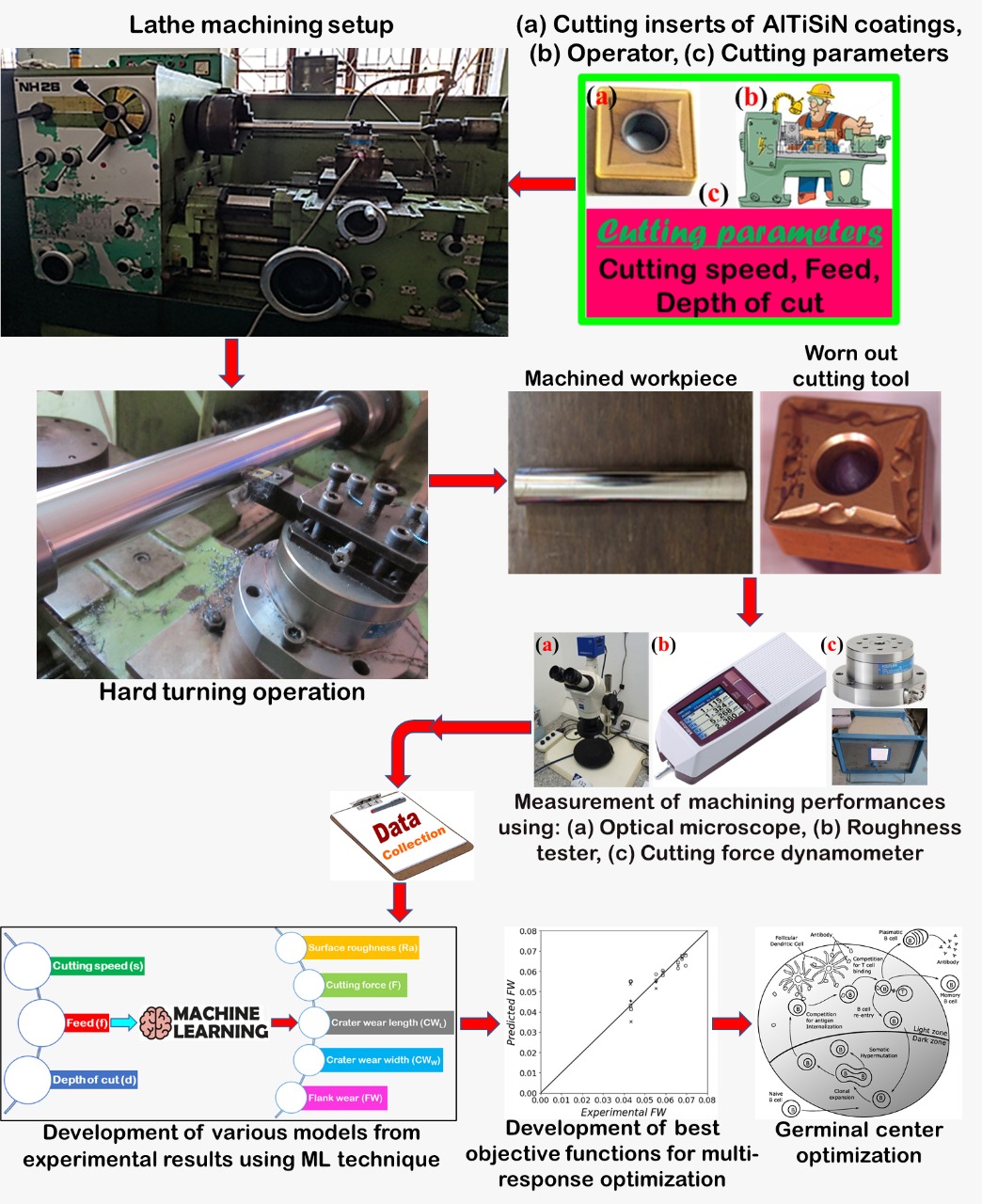}
\caption {The detailed framework used in optimization of machining parameters. \label{fig:chart}}
\end{figure}

Although various works related to the application of RSM and ML techniques in modeling and predictions of manufacturing processes are available in the literature, there is very limited research published related to optimization of process parameters of turning operation using ML-assisted meta-heuristic optimization algorithms. In this article, the different output responses (such as surface roughness, cutting force, crater wear length, crater wear width, and flank wear) during hard turning of AISI D6 steel are modeled in terms of speed, feed, and depth of cut. We have used four different ML algorithms for developing surrogate models to predict the output responses. The four ML algorithms are polynomial regression (PR), Random Forests (RF), Adaptive Boosting (AB), and Gradient Boosting (GB). The predictive capability of the ML models is assessed through a comparison of the different performance parameters such as mean squared error (MSE), mean absolute error (MAE), and the calculated value of coefficient of determination  $R^2$. The surrogate model learned through PR for five different output parameters is used to prepare a complex objective function and its optimization for finding the optimal set of process parameters that minimizes the outputs using system-based germinal center optimization algorithm.The detailed framework used in study is shown in figure \ref{fig:chart}. 

The rest of the article is arranged as follows: In section 2, the details of the experimental procedure for the generation of data to be used in ML-based modeling are discussed. In section 3, the coating characterisation of the cutting tool is discussed. In section 4, 5, 6 and 7 the effect of input parameters on the output responses are discussed. In section 8, 
the dataset used in the ML model development is discussed in detail, including the features and levels used and the scaling procedure adopted. The details of ML models are discussed in section 9. In section 10, the theory of the germinal center algorithm is discussed. In section 11, the ML-based prediction of responses and multi-objective optimization are discussed and finally, in section 12, the concluding remarks are presented.

\begin{table*}
\begin{center}
\begin{tabular} {c c c c c c c c c} 
  \toprule
     \hline
  \bfseries Element &  \multicolumn{1}{c}{\bfseries $Fe$} & \multicolumn{1}{c}{\bfseries $C$} & \multicolumn{1}{c}{\bfseries $Si$} & \multicolumn{1}{c}{\bfseries $Mn$} &
  \multicolumn{1}{c}{\bfseries $P$} & \multicolumn{1}{c}{\bfseries $S$} & \multicolumn{1}{c}{\bfseries $Cr$} & \multicolumn{1}{c}{\bfseries $W$}\\
  \midrule
$\%$ Wt. &  84.789 &  2.014 & 0.29  & 0.43 &  0.022 &  0.005 & 11.77  & 0.68
\end{tabular}
\caption{Chemical composition of AISI $D_6$ alloy die steel}\label{t_chemical}
\end{center}
\end{table*}
\section{Experimental setup of the turning operation and procedure of data collection}
The workpiece considered in this investigation corresponds to a shape of a cylindrical bar having a specification of $350$ mm length and $50$ mm diameter. The designation of the workpiece was AISI D6 steel. The particular grade of steel was selected because this steel grade of steel has vast applications in particularly tool and die making industries. And it has good dimensional stability with good wear and abrasion resistance. Table 1 shows the chemical composition of the material. As the machining is hard machining, to retain the hardness, the workpiece was undergone heat treatment, called tempering and quenching. Quenching was accomplished at 900$^\circ$ C and tempering was accomplished at 420$^\circ$ C. Finally, 65HRC was achieved. Uncoated carbide inserts are procured from Kennametal with a square shape having geometry SNMG 120408. Then the uncoated inserts are coated using S3P technology called scalable pulsed power plasma. The thin hard film coating which was deposited on the substrate was AlTiSiN coating. Before coating, carbide inserts are properly cleaned and polished. Cleaning was done using an ultrasonic bath and micro blasting. The scalable pulsed power plasma technique is a combined methodology of magnetron sputtering and cathode arc evaporation. The coating was done in Oreilikon Balzer, Pune, India. The Oerlikon Balzer make coating machine having model  INLENIA  was employed for this scalable pulsed power plasma coating. For coating, normally three operations are performed called, etching, heating and coating. In the beginning stage of the coating, the temperature of the coating chamber was maintained at 450$^\circ$C
and the vacuum pressure was maintained at $10^{-6}$ mill bar. The inserts are placed in the holder then it was rotated with a speed of 2 rpm speed.  For the smooth deposition of coating on the substrate with nano layer structure and uniform thickness, two fold motion was carried out. Two fold motion was accomplished with motor. Both etching and thermal treatment on the substrate are done. The etching was named as sputter etched. Sputter etching and thermal treatment are done to remove the surface contaminants from the substrate.  A voltage of 450V and current of 20 A is normally maintained. The coating was accomplished with the argon gas pressure of 350 MPa. Temperature was reduced gradually to 150℃ after completion of the coating. Then finally the inserts with coating was kept outside with natural air cooling. The AlTiSiN coating deposition was completed using one target each of AlTiN and TiSiN with 5 KW target power and 99.95\% purity. The thickness of AlTiN coating was 0.5± 0.1 micron and that of TiSiN was 0.2 ± 0.1 micron. The final coating layer thickness was 1.3 ± 0.1 micron. Good adhesion strength was observed between the coating material and substrate. Inserts (substrate) are placed in a negatively biased coating chamber and from the positively charged source (AlTiSiN), arc was produced. Then positively charged ions are attracted towards the substrate and got deposited. After coating, the inserts are clamped on a tool holder called PSBNR 2020K12 having an approach angle $75^o$ and nose radius of 0.8 mm. 
The dry hard turning on AISI D6 steel was performed on a high precision heavy duty lathe (Model NH-26, make- HMT). The maximum spindle power was 11KW and the spindle speed was in the range between 40-2040 rpm. For cutting force (Fz) measurement, 4- component piezoelectric dynamometer was used, (Make- Kistler, Model-9272), The Mitutoyo SJ-210 roughness tester was employed to measure the machined surface roughness. Both optical and scanning electron microscope was employed to analyze the machined surface morphology and tool wear (both at flank and rake face). In the current experimental investigation, three input parameters are considered such as speed, feed, and depth of cut. As the depth of cut is not an effective input variable compared to the speed and feed only four levels are considered for depth of cut, whereas for both speed and feed seven levels were chosen. Speed (40, 50, 55, 60, 70, 80 and 90 m/min), feed (0.04, 0.06, 0.08, 0.1, 0.12, 0.14 and 0.16 mm/rev) and depth of cut (0.2, 0.3, 0.4 and 0.5mm) to analyze  different machining characteristics in dry cutting condition. The above parameters and their levels are selected according to the literature review and the tool manufacturer’s recommendation. The detail procedure adopted in the present work with optimization is shown in Fig.\ref{fig:chart}.

\section{Coating characterization of cutting tool}
\begin{figure}
\centering
 \includegraphics[width=0.8\textwidth]{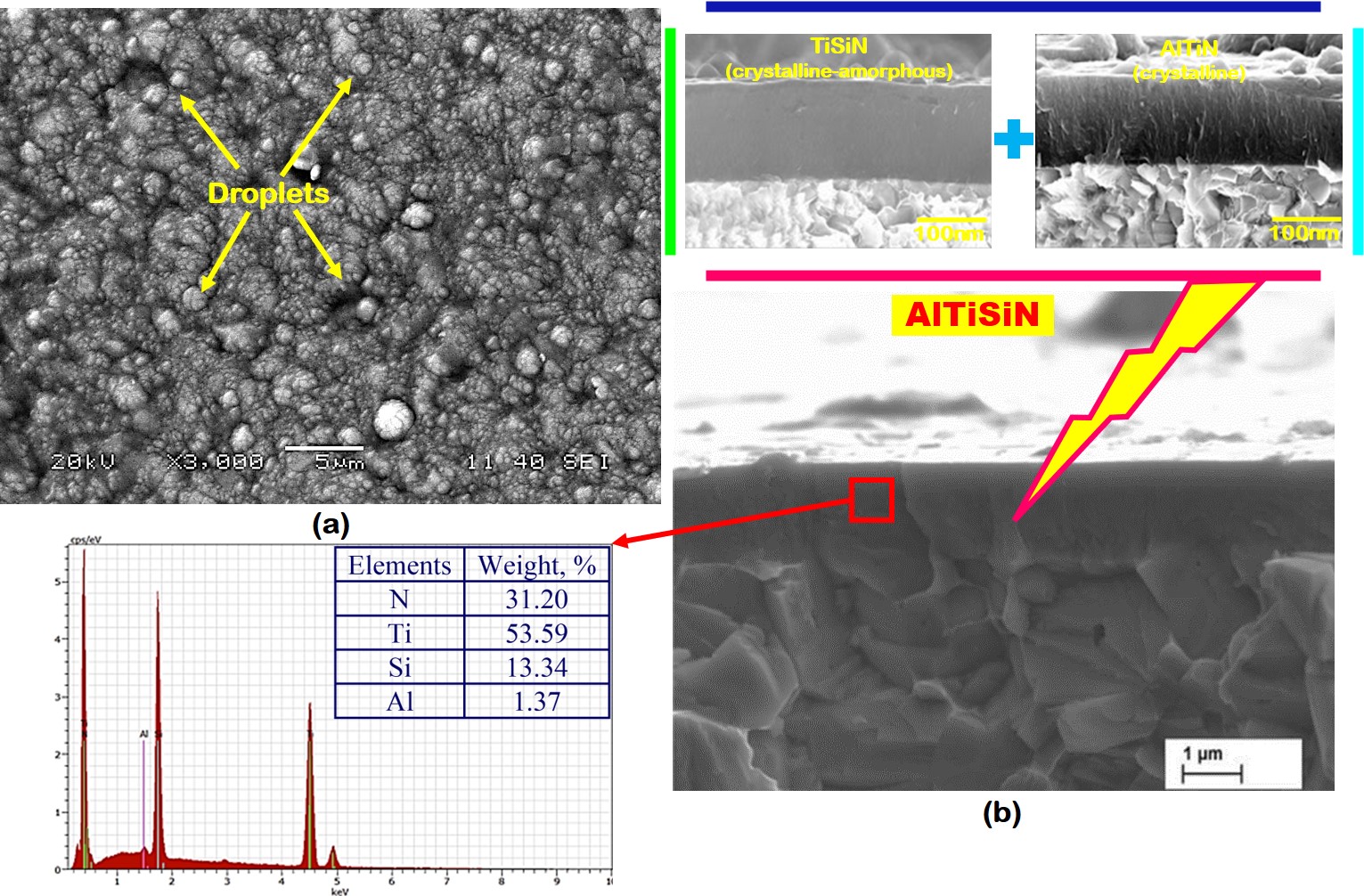}
 
 \caption{Figure  SEM images showing (a) surface morphology and (b) cross-sectional morphology of AlTiSiN coatings with EDS\label{fig:SEM}}
\end{figure}

If machining will be conducted with coated inserts, in dry cutting circumstances, coating should possess two important characteristics like anti- wear and anti-oxidant. The surface morphology of AlTiSiN coated carbide insert is shown in Figure \ref{fig:SEM}. The scanning electron microscopic  (SEM) image (figure \ref{fig:SEM}) delineated that high dense structure with droplet deposition was found. The cross section of AlTiSiN is illustrated in Fig. columnar free structure was observed from the image. This was mainly due to the existence of crystalline-amorphous phases of both AlTiN and TiSiN structures. Nano crystalline TiN is dispersed to large amount of Si3N4 amorphous material in TiSiN resulting a smoother surface with glassy appearance. TiSiN offers high oxidant resistant because $SiO_2$ forms an antioxidant layer resulting higher hardness. The hardness mainly depends upon the following factors, 
a) robust covalent bond formation like SiN compound, 
b) Strengthening the structure by incorporating various atoms and c) Grain growth of nitrides due to strong amorphous surrounded SiN compounds in grain refinement process.

Also higher hardness will be achieved with higher silicon content and lower crystalline size. Moreover, the higher volume of silicon content and smaller crystalline size is responsible for the evanescement of columnar structure.

\section{Analysis of tool flank wear}
\begin{figure}
\centering
 \includegraphics[width=1\textwidth]{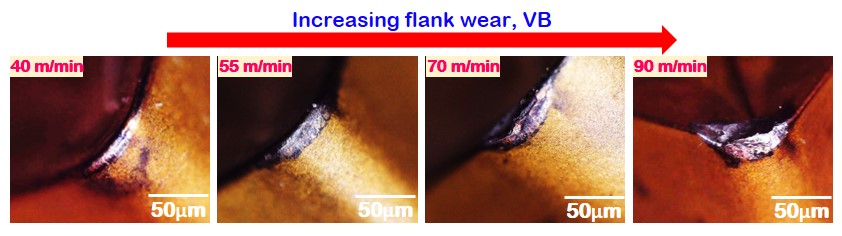}
 \caption{Optical micro-graphs illustrating the flank wear under different cutting speeds at f = 0.16 mm/rev and d = 0.5 mm\label{fig:optical1}}
\end{figure}

\begin{figure}
\centering
 \includegraphics[width=0.98\textwidth]{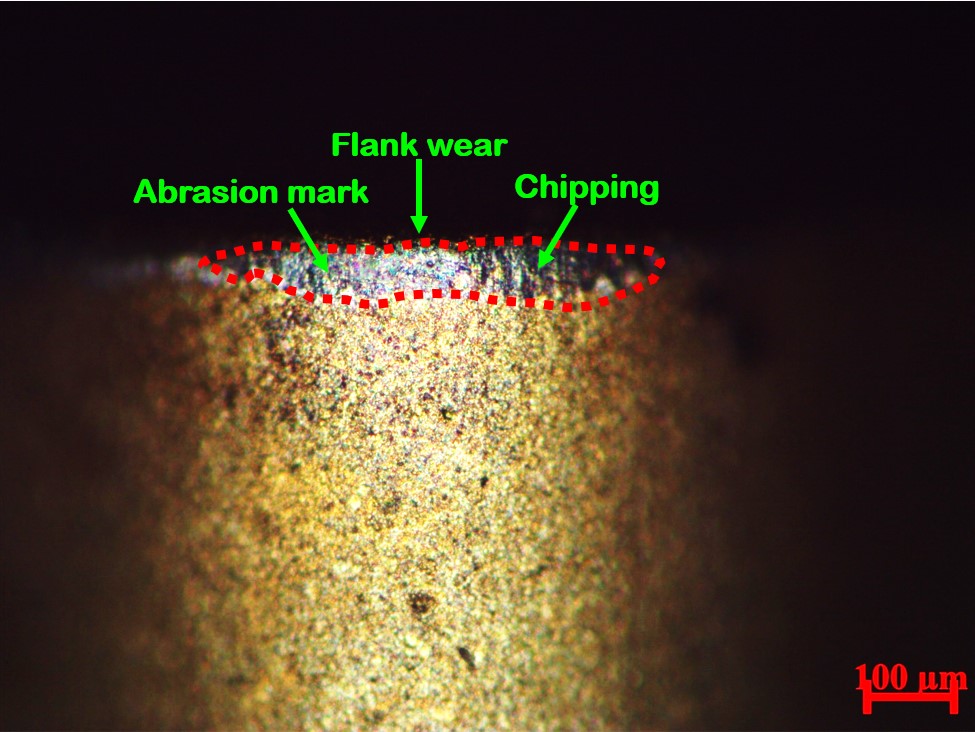}
 \caption{Microscopic view of the flank surface of AlTiSiN coated insert \label{fig:micro2}}
\end{figure}

\begin{figure}
\centering
 \includegraphics[width=0.98\textwidth]{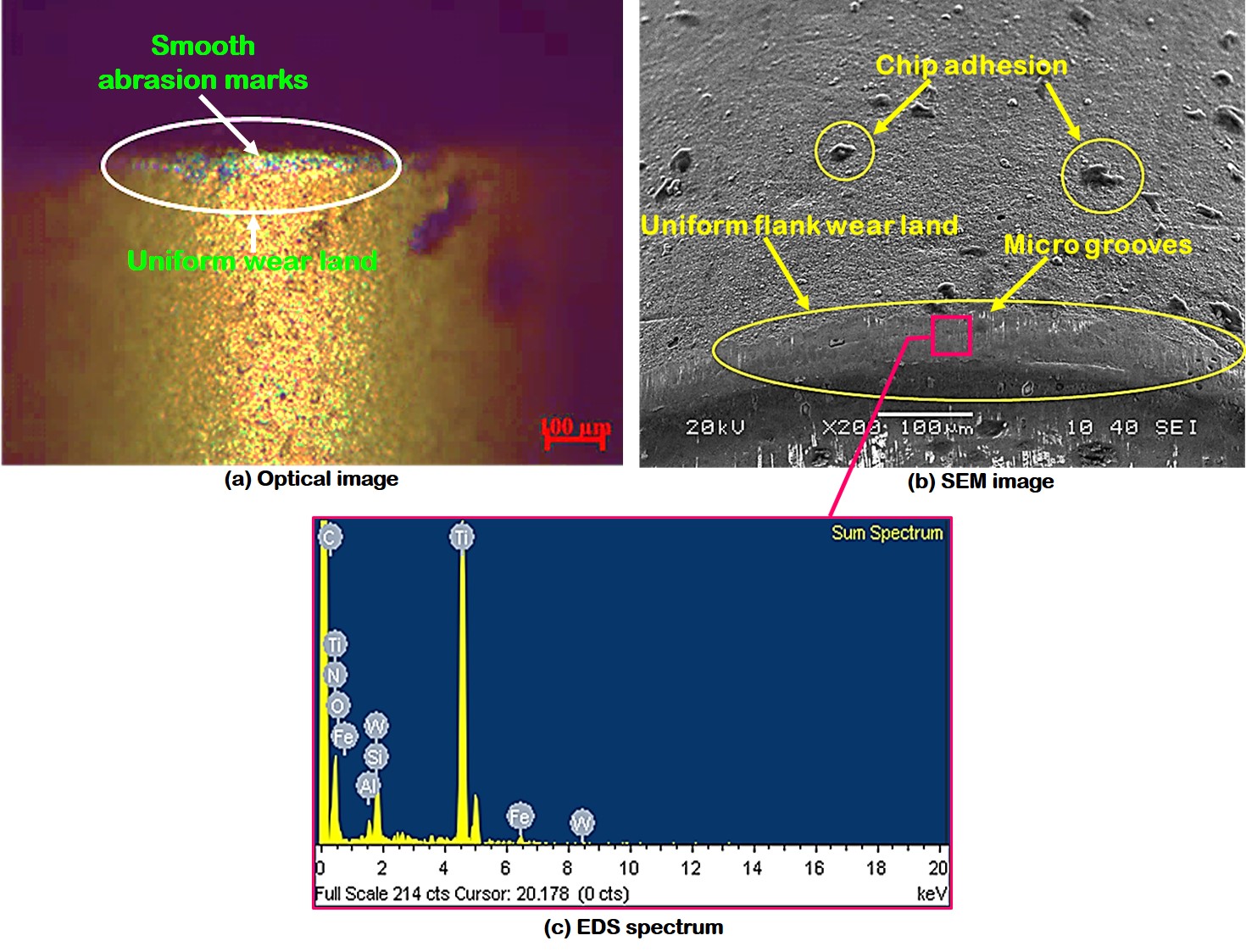}
 \caption{Flank wear evaluation of SPPP-AlTiSiN coated carbide tool under dry cutting conditions at v = 70 m/min, d = 0.4 mm, f = 0.04 mm/rev\label{fig:flank1}}
\end{figure}

As shown in figure \ref{fig:optical1}, it was evident that flank wear is increasing with the cutting speeds, four cutting speeds are chosen and feed and depth of cut were kept constant as 0.16mm/rev and 0.5mm respectively. The increment  trend of flank wear  might be due to the thermal softening near the cutting edge of the insert. Thermal softening is mainly influenced by the temperature development or heat generation at tool-work interface during machining. Moreover, heat is an energy normally flows from high temperature region to low temperature region, but it needs some time. But when machining will be accomplished at higher cutting speed, the time will be insufficient for effective heat transfer. Thermal conductivity of the material is influenced by its hardness. As the work piece was heat treated for the hardness enhancement, its thermal conductivity might be hampered. That’s why chances for effective heat transfer to the work piece will be less. As a result, inserts might be exposed to high temperature resulting thermal softening near the cutting edge. For which more wear at the flank surface of the insert was observed at higher speed values. As stated in the optical microscopic images in Fig.\ref{fig:optical1} Uniform wear land and abrasion marks are observed. Smooth abrasion marks are observed when machining was accomplished at low and medium speed shown in Fig.\ref{fig:flank1}. But when machining was done with high speed, thick abrasion, chipping and plastic deformation observed which is shown in Fig.\ref{fig:micro2}. Adhesion of chips and micro grooves are also observed which is shown in the SEM image (Fig.\ref{fig:flank1}). The abrasion marks are observed due to abrasive nature of coarse grain structures and different carbides precipitation, which might be developed due to heat treatment of the test specimen for enhancement in hardness. Two main wear mechanisms are observed adhesion and abrasion. Both the mechanisms are normally observed when continuous machining is conducted for a long time. Because due to continuous machining, insert is subjected to rapid cyclic loading. 

\section{Analysis of cutting force}
\begin{figure}
\centering
\includegraphics[width=0.8\textwidth]{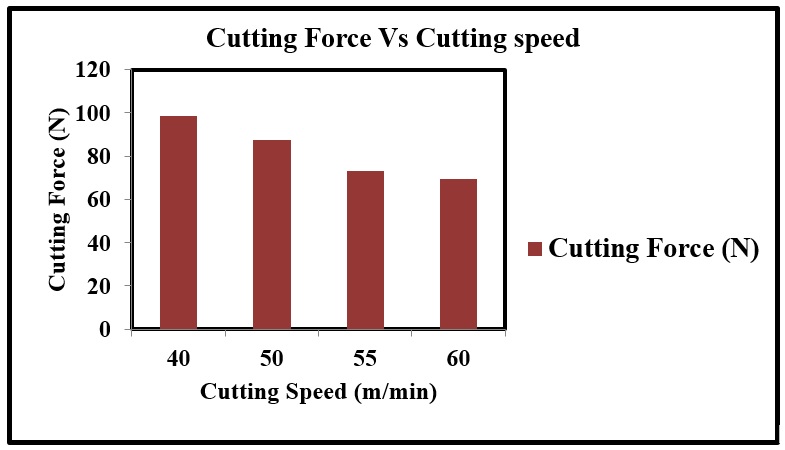}\\
 
 \caption{Cutting force vs. cutting speed (low and medium range)\label{fig:cf_s_lmr}}
\end{figure}

\begin{figure}
\centering

\includegraphics[width=0.8\textwidth]{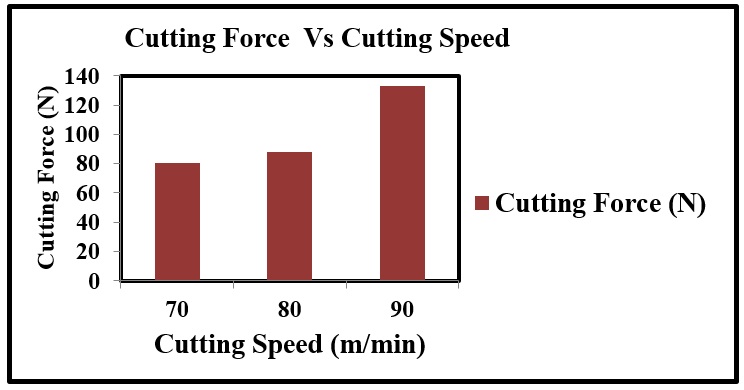}
 
 \caption{Cutting force vs. cutting speed (higher range of speed)\label{fig:cf_s_hsr}}
\end{figure}

\begin{figure}
\centering
\includegraphics[width=0.8\textwidth]{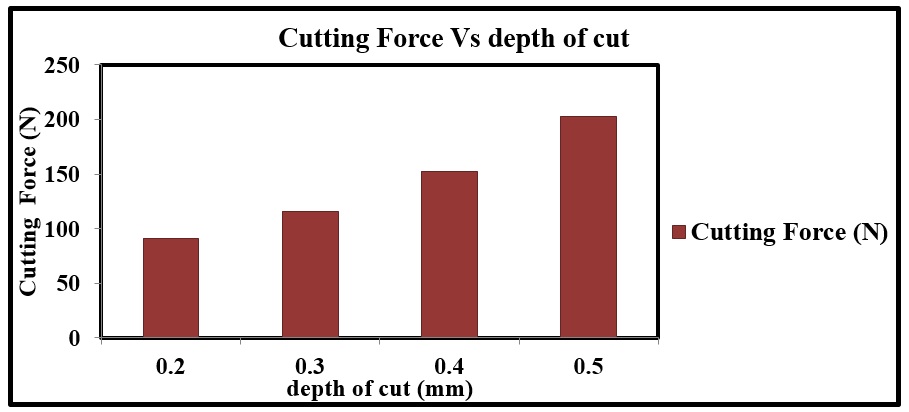}
 \caption{Cutting force vs. depth of cut\label{fig:cf_d}}
\end{figure}

Out of different machining characteristics, cutting force is one of the most important characteristics. Power consumption, dimensional deviation and tool life were significantly affected by the cutting force. Cutting force was measured with the help of dynamometer. Experiments are carried out in seven different speeds (40, 50, 55, 60, 70, 80 and 90m/min). From the graphical representation shown in Fig. \ref{fig:cf_s_lmr}, it was observed that at four different levels of speed such as 40, 50, 55 and 60m/min, the decrements in the cutting force was observed. This might be due to the thermal softening of the work piece. Because at high speed machining, time will be limited for the effective heat transfer to the surroundings from the machining zones. So there is a chance of more heat transfer to the work piece. Moreover, at high speed, more temperature is generated due to tool-work and tool-chip friction. Due to which shear strength of the specimen diminished and grain boundary disruption occurred resulting depletion of cutting force. As discussed earlier (in the section Analysis of tool flank wear), tool wear was highly influenced by the cutting speed. Particularly for three ranges of cutting speed i.e., 70, 80 and 90m/min, rapid wear on the flank surface was observed. That’s why there was a rapid increment in the cutting force as shown in Fig.\ref{fig:cf_s_hsr}. Cutting force increment was also observed with the depth of cut shown in Fig.\ref{fig:cf_d} this might be due to the increment of shear plane area, chip thickness and material removal rate. Moreover, similar trend was observed for cutting force with axial feed rates as shown in Fig.\ref{fig:effect}. With lower feed values, less force was observed and with higher feed values, more force was observed. This might be due to the decrements of contact length between tool tip and test specimen, for which sharpness of the tool retained resulting less force. But as feed increased, contact length increased between tool tip and work piece for which sharpness of the cutting edge demolished resulting rough machining and generation of more cutting force.

\begin{figure}
\centering
\includegraphics[width=0.8\textwidth]{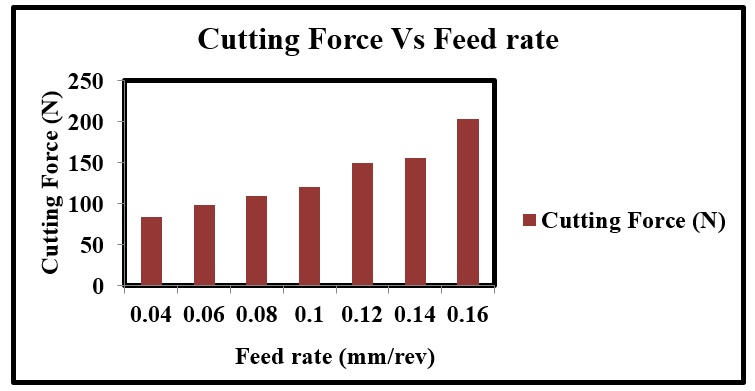}
 \caption{Cutting force vs. feed rate\label{fig:cf_f}}
\end{figure}

\section{Analysis of surface roughness}

\begin{figure}
\centering
 \includegraphics[width=0.8\textwidth]{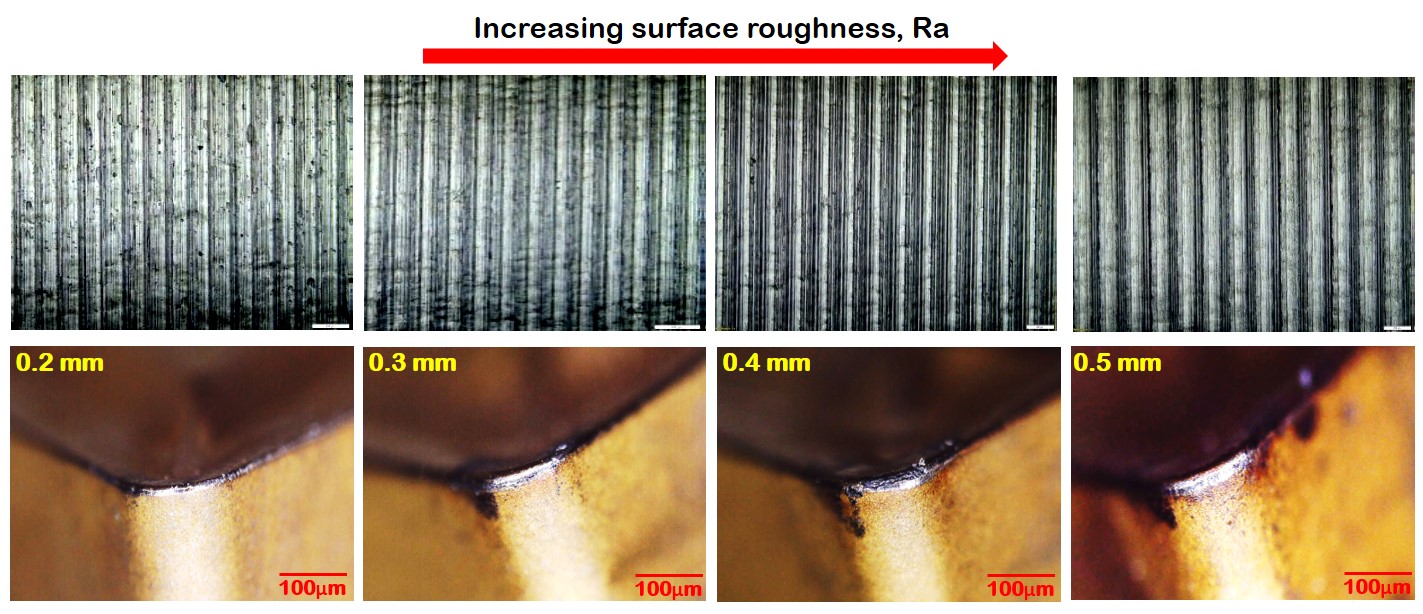}
 \caption{Effect of flank wear on machined surface morphology under varying depth of cut at cutting speed = 40 m/min, feed = 0.16 mm/rev \label{fig:effect}}
\end{figure}

\begin{figure}
\centering
 \subfloat[]{\includegraphics[width=0.8\textwidth]{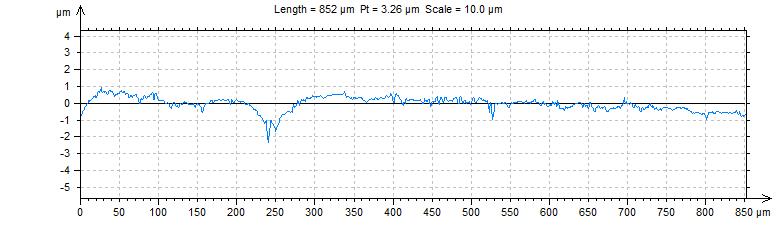}}\\
 \subfloat[]{\includegraphics[width=0.8\textwidth]{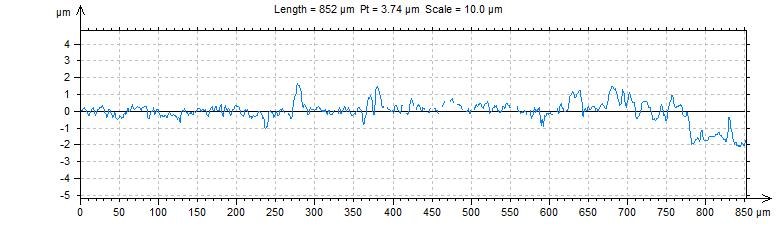}}\\
 \subfloat[]{\includegraphics[width=0.8\textwidth]{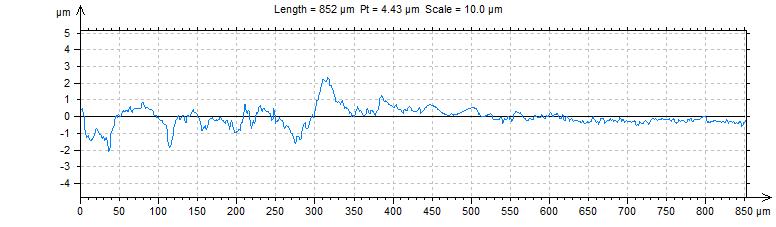}}\\
 \subfloat[]{\includegraphics[width=0.8\textwidth]{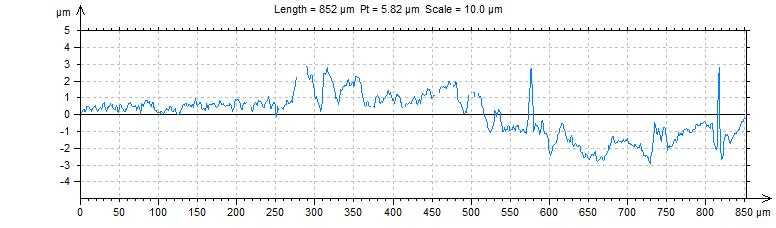}}\\
 \caption{Roughness profile with different depth of cut a) 0.2mm b) 0.3mm c) 0.4mm d) 0.5mm\label{fig:roughnessprofile}}
\end{figure}

From the figure \ref{fig:effect} it was found that with the increment in the depth of cut, surface roughness increased by keeping speed and feed constant. This might be due to the increment of the contact area between the tool and work piece at the machining zone resulting extreme ploughing force. That leads to the side flow of material which is responsible for the machined surface degradation. Another reason might be chatter or self-induced vibration due to tool wear. Moreover, more elastic deformation and squeezing of materials was observed at tool-work contact zone in lower value of depth of cut compared to higher value of  depth of cut. That’s why more plasticized material flow might be occurred through insert’s side way resulting machined surface deterioration. Self-induced chatter or vibration is more susceptible in machining of hard materials. In current experimental investigation, material having hardness 65 HRC was used. So chatter was noticed. Moreover, with the increment in depth of cut, contact region between tool flank face and work piece surface increased. For which both temperature and pressure might be increased at the contact zone resulting severe abrasion, burnt marks, plastic deformation at the cutting edge was observed. That’s why deeper and wider feed marks on the machined surface were observed as noticed in the figure. The roughness profiles generated by surface roughness tester with varying depth of cut conditions are shown in figure \ref{fig:roughnessprofile} below where speed and feed were kept constant as 40 m/min and 0.16 mm/rev respectively. 

\section{Effect of chip-tool contact length on crater wear}

\begin{figure}
\centering
 \includegraphics[width=0.98\textwidth]{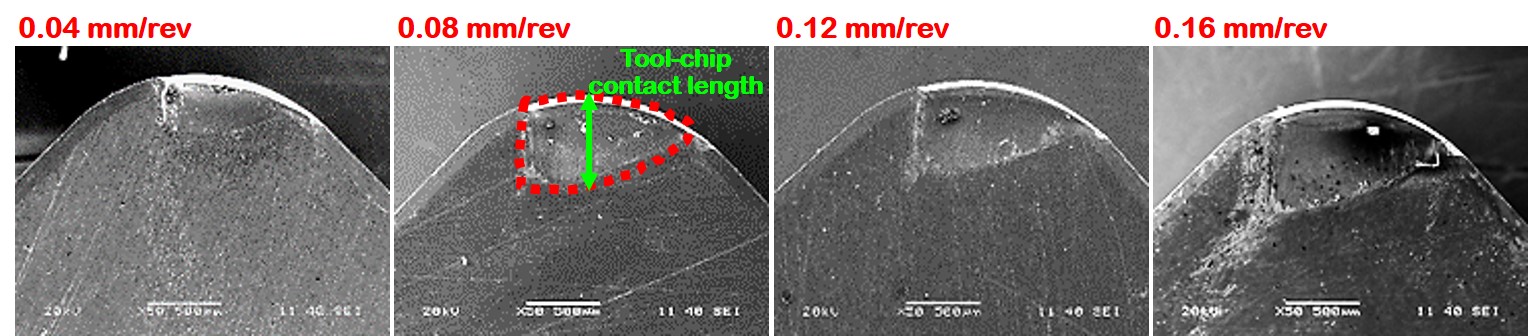}
 \caption{Crater wear due to tool-chip sliding contact length with varying feed at v= 90 m/min; d= 0.2 mm \label{fig:crater}}
\end{figure}

When machining was conducted at higher feed values continuous spiral chips are produced. But one interesting phenomena was observed that chips were up curl in nature. There was a contact between the chips and the work piece at their end and chips rolling were observed for more than one revolution. Straightened chips with increased curvature radius are observed. This might be due to the end contact of the chips with the specimen. The longitudinal grooves are observed due to the chip sliding at the tool-chip interface. In the contact region, the grooves are observed but with the chip flow direction. Due to the continuous chip sliding on the rake surface an irregular side flow of chips is observed resulting a long tool-chip contact length. Moreover, more wear on the rake surface was observed. From the Fig.\ref{fig:crater}, it was clear that crater wear increased with the feed rate because of increment in the tool-chip contact length. Lowest tool wear was observed with low speed and feed. The microscopic view of the rake surface of the AlTiSiN coated insert was shown in Fig.13. Various characteristics are observed like abrasion marks, chipping, multi hot spots and burn spots. With high speed and feed, these characteristics are observed. Both chip sticking and chip sliding were the influencing factors for the above said characteristics. Abrasion marks were mainly observed due to the chip sliding. But hot spots, burn spots and chipping might be influenced through chip sticking. Due to machining a hard material, hot and hard chips are produced. As discussed in the above section, more heat might be transferred to tool material compared to work piece material. So heat will be retained on the tool surface. So if in this situation, hot and hard chips will pass, due to high temperature, chip welding or sticking will be occurred resulting hot spots or burn spots. And chip- ping was observed on the rake surface due to collision of hot and hard chips with the rake face of the insert.
\begin{figure}
\centering
 \includegraphics[width=0.98\textwidth]{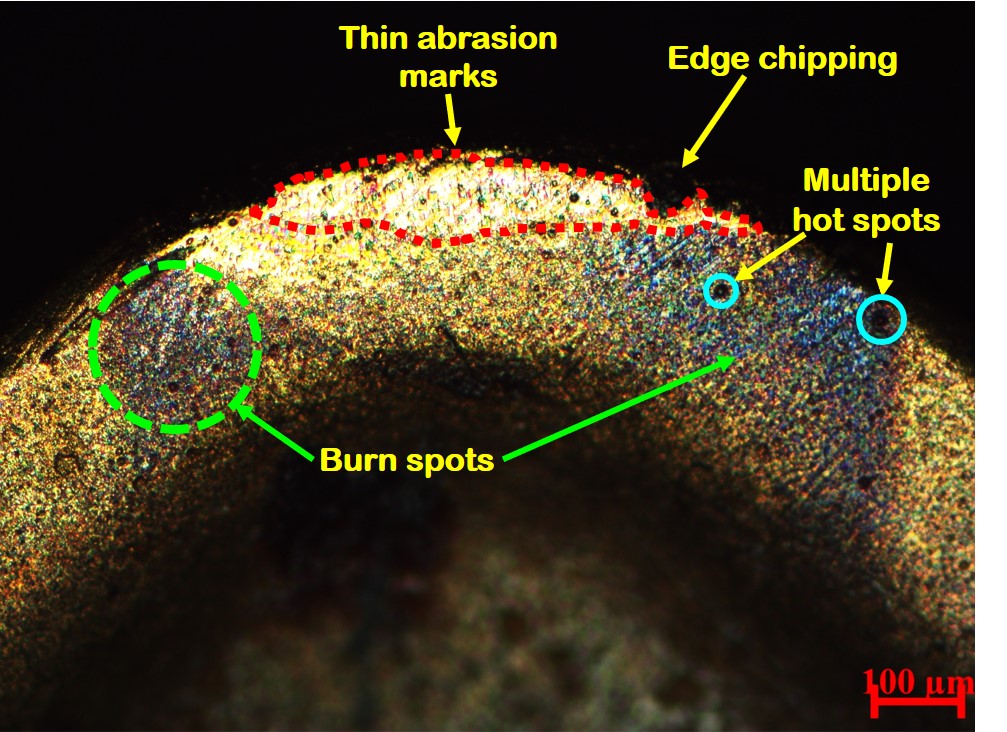}
 \caption{Microscopic view of the rake surface of AlTiSiN coated insert \label{fig:micro1}}
\end{figure}

 

 

\section{Description of the dataset used in ML model development}
The dataset used in this article has eight columns as shown in the table. \ref{tbl:data}. In this tabular representation, each column represents a characteristic attribute and each row represents a specific sample. The first three columns, s, f, d are used as input to the ML models. These represent features that the algorithm uses to model the target output. The next five columns are the outputs of the ML models. The total number of samples is 49. The input features were normalized to the range -1 to 1 to ensure efficient gradient-based learning. This min-max scaling method for normalization can be defined as follows:      
\begin{equation}
\phi_s=\frac{\phi-\phi_{min}}{\phi_{max}-\phi_{min}} 
\end{equation}
where, $\phi_s$ is the new scaled data, $\phi$ is the original cell data, $\phi_{min}$ and $\phi_{max}$ are the minimum and maximum value in corresponding column. The importance of scaling of input data is to make the learning process numerically stable. The inputs/ features of the ML models are s, f and d and the outputs/ levels are $Ra$, $F$, $CW_L$, $CW_W$ and $FW$. We have prepared two train-test datasets ($D_1$ and $D_2$) for the development of regression models. The training dataset of $D_1$ consist of all the data as presented in table \ref{tbl:data}, except sl. no 8-14, those correspond to 50 m/min speed. The data correspond to 50 m/min was used for testing of the developed ML regression models. The train-test dataset $D_2$ consist of arbitrary testing data from various speed from table \ref{tbl:data}. Those are (40, 0.04, 0.2), (50, 0.06, 0.2), (55, 0.16, 0.4), (60, 0.14, 0.5), (70, 0.12, 0.2), (80, 0.1, 0.5) and (90, 0.06, 0.4) in the order (s, f, d). The dataset except the following combinations were using in model training.

\section{Machine learning models for predictive modeling of machining parameters}
\label{MLmodels}
Machine Learning is a sub-field of AI that has its focus on enabling computer-based systems to develop predictive models purely from data. The learning aspect in the name represents the ability of such algorithms to improve their performance at a task by processing data, without needing any updated human or domain expert instructions. The data used in ML occurs in pairs of features and targets in supervised learning. By use of machine learning, a function relating the inputs features to the target outputs can be developed by using suitable datasets. There are different ML algorithms available through which a surrogate model can be developed for a set of input and output data, those are linear regression (LR), polynomial regression (PR), support vector regression (SVR), and different tree-based algorithms such as Adaptive Boosting (AdaBoost), Gradient-Boost (GB) and Random Forests (RF). The linear regression (LR) is suitable for simple type linear relationships and is a subset of polynomial regression. In this investigation,  polynomial regression, Random Forests, Gradient Boosting, and Adaptive Boosting algorithms for predictive modeling of the machining parameters is considered. Fig. 14 shows the machine learning frame work used for the modeling of input variables . The decision tree structure used in tree based ML algorithm is shown in Fig. 15. The predictive performance of the above-mentioned ML algorithms was assessed by using $R^2$, $MSE$ (mean absolute error), and $MAE$ (mean absolute error). Those can be defined as follows:        

\label{t1}
\begin{center}
\setlength\extrarowheight{-1pt}
\begin{tabular}
 {||c c c c c c c c c||} 
 \hline
 sl. no. & $s$ & $f$ & $d$ & $Ra$ & $F$ & $CW_L$ & $CW_W$ & $FW$\\
 [0.2ex] 
 \hline\hline
 1 & 40 & 0.04  & 0.2 & 0.45 & 84 & 0.25 & 0.036 & 0.044\\ 
 \hline
 2 & 40 & 0.06  & 0.2 & 0.5 & 98.4 & 0.3 & 0.053 & 0.044\\ 
 \hline
 3 & 40 & 0.08  & 0.3 & 0.5 & 110 & 0.41 & 0.075 & 0.057\\ 
 \hline
 4 & 40 & 0.1  & 0.3 & 0.65 & 120.77 & 0.45 & 0.089 & 0.060\\ 
 \hline
 5 & 40 & 0.12  & 0.4 & 0.74 & 149.99 & 0.56 & 0.10 & 0.068\\ 
 \hline
 6 & 40 & 0.14  & 0.4 & 0.80 & 155.63& 0.57 & 0.12 & 0.071\\ 
 \hline
 7 & 40 & 0.16  & 0.5 & 0.89 & 202.99 & 0.699 & 0.159 & 0.070\\ 
 \hline
 8 & 50 & 0.04  & 0.2 & 0.38 & 72.85 & 0.23 & 0.032 & 0.043\\ 
 \hline
 9 & 50 & 0.06  & 0.2 & 0.43 & 87.55 & 0.29 & 0.048 & 0.043\\
 \hline
 10 & 50 & 0.08  & 0.3 & 0.51 & 95.65 & 0.377 & 0.07 & 0.055\\
 \hline
 11 & 50 & 0.1  & 0.3 & 0.58 & 107.52 & 0.42 & 0.084 & 0.058\\ 
 \hline
 12 & 50 & 0.12  & 0.4 & 0.67 & 137.15 & 0.52 & 0.10 & 0.065\\
 \hline
 13 & 50 & 0.14  & 0.4 & 0.74 & 141.91 & 0.54 & 0.122 & 0.069\\
 \hline
 14 & 50 & 0.16 & 0.5 & 0.83 & 189.38 & 0.66 & 0.15 & 0.067\\ 
 \hline
 15 & 55 & 0.04  & 0.3 & 0.35 & 54.99 & 0.23 & 0.034 & 0.05\\ 
 \hline
 16 & 55 & 0.06  & 0.3 & 0.40 & 73.15 & 0.30 & 0.05 & 0.05\\ 
 \hline
 17 & 55 & 0.08  & 0.2 & 0.46 & 96.99 & 0.32 & 0.06 & 0.044\\ 
 \hline
 18 & 55 & 0.1  & 0.2 & 0.53 & 106.08 & 0.36 & 0.07 & 0.05\\ 
 \hline
 19 & 55 & 0.12 & 0.5 & 0.68 & 174 & 0.62 &0.12 & 0.06\\ 
 \hline
 20 & 55 & 0.14  & 0.5 & 0.74 &180.88 & 0.64 & 0.14 & 0.06\\ 
 \hline
 21 & 55 & 0.16 & 0.4 & 0.76 & 137 & 0.54 & 0.13 & 0.07\\ 
 \hline
22 & 60 & 0.04  & 0.3 & 0.33 & 50.9 & 0.23 & 0.03 & 0.05\\ 
 \hline
23 & 60 & 0.06  & 0.3 & 0.38 & 69.43 & 0.3 & 0.05 & 0.05\\ 
 \hline
24 & 60 & 0.08  & 0.2 & 0.45 & 94.4 & 0.32 & 0.06 & 0.05\\ 
 \hline
25 & 60 & 0.1  & 0.2 & 0.51 & 103.59 & 0.35 & 0.07 & 0.05\\ 
 \hline
26 & 60 & 0.12  & 0.5 & 0.51 & 170.04 & 0.61 & 0.12 & 0.06\\ 
 \hline
27 & 60 & 0.14  & 0.5 & 0.51 & 175.62 & 0.63 & 0.13 & 0.06\\ 
 \hline
28 & 60 & 0.16  & 0.4 & 0.51 & 132.85 & 0.53 & 0.13 & 0.07\\ 
 \hline
29 & 70 & 0.04  & 0.4 & 0.51 & 57 & 0.28 & 0.04 & 0.06\\ 
 \hline
30 & 70 & 0.06  & 0.4 & 0.51 & 80.18 & 0.35 & 0.05 & 0.06\\
 \hline
31 & 70 & 0.08  & 0.5 & 0.51 & 136 & 0.52 & 0.09 & 0.06\\
 \hline
32 & 70 & 0.01  & 0.5 & 0.51 & 154.53 & 0.57 & 0.1 & 0.06\\
 \hline
33 & 70 & 0.12  & 0.2 & 0.51 & 108 & 0.36 & 0.06 & 0.06\\
 \hline
34 & 70 & 0.14  & 0.2 & 0.51 & 108.46 & 0.37 & 0.07 & 0.06\\
 \hline
35 & 70 & 0.16  & 0.3 & 0.51 & 109 & 0.45 & 0.1 & 0.07\\
 \hline
36 & 80 & 0.04  & 0.4 & 0.51 & 64.48 &  0.31 & 0.03 & 0.06\\
 \hline
37 & 80 & 0.06  & 0.4 & 0.51 & 88.01 & 0.38 & 0.04 & 0.06\\
 \hline
38 & 80 & 0.08 & 0.5 & 0.51 & 137.31 & 0.53 & 0.08 & 0.06\\
 \hline
39 & 80 & 0.1  & 0.5 & 0.51 & 155.03 & 0.59 & 0.09 & 0.06\\
 \hline
40 & 80 & 0.12  & 0.2 & 0.51 & 114.61 & 0.38 & 0.05 & 0.06\\
\hline
41 & 80 & 0.14  & 0.2 & 0.51 & 115.55 & 0.39 & 0.06 & 0.07\\
\hline
42 & 80 & 0.16  & 0.3 & 0.51 & 119 & 0.48 & 0.09 & 0.08\\
\hline
43 & 90 & 0.04  & 0.5 & 0.51 & 96 & 0.42 & 0.04 & 0.07\\
\hline
44 & 90 & 0.06  & 0.5 & 0.51 & 118.56 & 0.49 & 0.05 & 0.07\\
\hline
45 & 90 & 0.08  & 0.4 & 0.51 & 101.07 & 0.42 & 0.04 & 0.07\\
\hline
47 & 90 & 0.06  & 0.4 & 0.51 & 133 & 0.5 & 0.05 & 0.07\\
\hline
48 & 90 & 0.14  & 0.3 & 0.51 & 134.31 & 0.52 & 0.06 & 0.08\\
\hline
49 & 90 & 0.16  & 0.2 & 0.51 & 127 & 0.46 & 0.08 & 0.08\\
\hline
\end{tabular}
\captionof{table}{Dataset used in ML model development}
\label{tbl:data}
\end{center}

\begin{equation}
R^2=1- \frac {\sum_i(\phi_i-\hat{\phi}_i)^2} {\sum_i(\phi_i-\bar{\phi_i})^2}
\end{equation}
\begin{equation}
MSE=\frac{1}{N} \sum_{i=1}^{N}(\phi_i-\hat{\phi}_i)^2
\end{equation}

\begin{equation}
MAE=\frac{1}{N} \sum_{i=1}^{N} |\phi_i-\hat{\phi}_i|.
\end{equation}
Here $|.|$ represents the absolute value.
\begin{figure}
\centering
\includegraphics[width=0.6\textwidth]{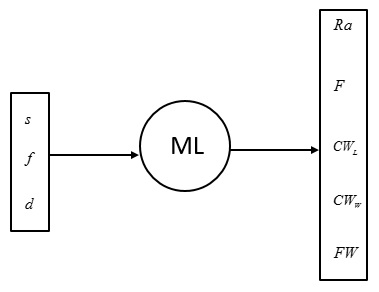}
\caption {The machine learning framework used in the modeling of machining parameters.\label{fig:1}}
\end{figure}

\begin{figure}
\centering
\includegraphics[width=0.4\textwidth]{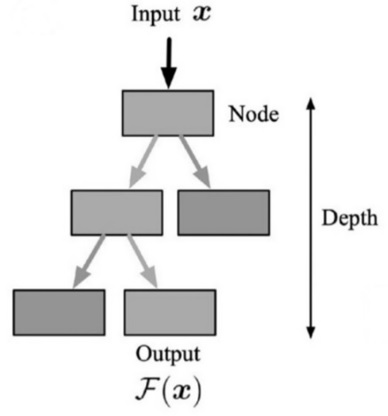}
\caption {The structure of a decision tree as used in tree based ML algorithms (RF, AB and GB)  \cite{panda2022machine}. \label{fig:tree}}
\end{figure}

\begin{figure*}
\centering
\subfloat[]{\includegraphics[width=0.5\textwidth]{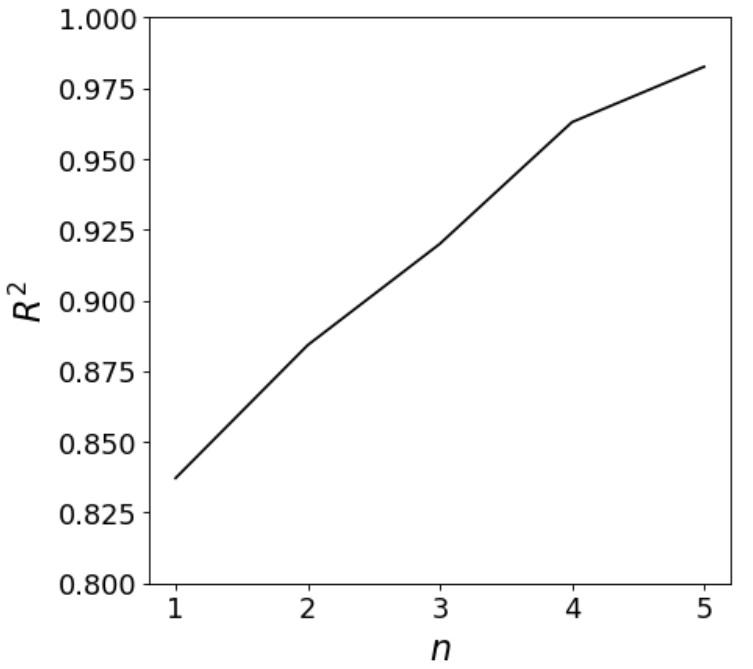}}
\subfloat[]{\includegraphics[width=0.5\textwidth]{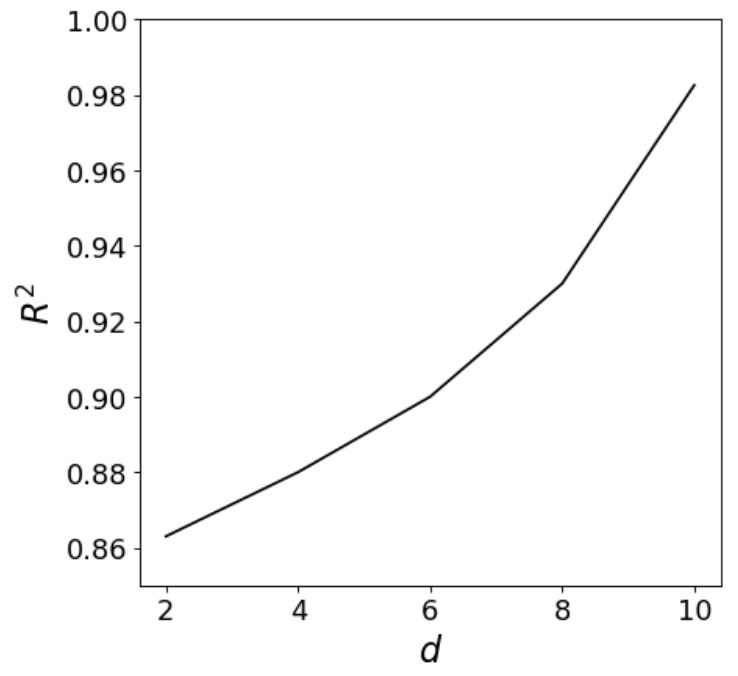}}
\caption {Hyperparameter optimization of random forest, for a) d=10 and for b) n=5. \label{fig:hp}}
\end{figure*}

\subsection{Polynomial regression (PR)}
A functional relationship between the features and targets in the data may be developed using arbitrary approaches. Algorithms such as deep neural networks and ensemble approaches lead to functional relationships that are very flexible. But these functions are often too complicated for human analysis a posteriori. This limits the interpretability of the ML model. For human interpretable models, we utilize polynomial regression. Polynomial regression represents the target as a linear combination of basis functions of the input features \citep{ostertagova2012modelling}. These basis functions can be arbitrarily convoluted based on the complexity of the problem. Based on experimentation with the dataset, we use the second-order polynomial basis function. The mathematical expression for the second-order polynomial regression model can be written as follows:
\begin{equation}
y=\beta_0+ \sum_{i=1}^{k}\beta_ix_i+\sum_{i=1}^{k}\beta_{ii} x_i^2+ \sum_{i}\sum_{j}\beta_{ij}x_ix_{j}+\epsilon
\end{equation}

The $\beta_0$ term represents the intercept and captures constant baseline effects that are independent of the input features. The $\beta_i$  and $\beta_{ii}$ terms capture the linear and quadratic relationships between the input features and the target value. The $\beta_{ij}$ term represents the interactions amongst the input features. Thus the coefficients of this polynomial closure have clear human interpretability. Training of this model invokes gradient descent to find the appropriate coefficients for the dataset. 

\subsection{Random Forests (RF)}
Random Forests are an ensemble learning algorithm where a meta-model is developed using a group (or ensemble) of decision trees \citep{panda2022machine,chung2021data}. Decision Trees (also referred to as Classification And Regression Trees) are a simpler modeling algorithm. Each decision tree represents a set of inequalities that leads to a partitioning of the feature space. Different subsets of this partitioning are assigned specific values of the target. A simple decision tree is shown in figure \ref{fig:tree}. While decision trees are simple to use, they lead to overfitting. To ameliorate this, an ensemble of decorrelated trees can be trained on the same data. Here the decorrelation is produced by bootstrapping the dataset and feature bagging. The final prediction of the Random Forest model is via aggregation of the predictions of the individual decision trees \citep{chung2020random}. Due to their flexibility and accuracy, Random Forests have been extensively used in physics and physical problems, such as renewable energy generation \citep{serras2019combining}, turbulence modeling \citep{heyse2021estimating}, Meteorological prediction \citep{loken2019postprocessing}, combustion simulations \citep{chung2021interpretable}, etc. The important hyperparameters of the random forest are the number of decision trees $(n)$ and maximum depth $(d)$. Hyperparameter optimization is an important task in any ML development. In this article, a manual approach was followed, in optimizing the n and d. The graphs of $R^2$ variation in terms of n and d are presented in figure \ref{fig:hp}. For figure \ref{fig:hp} a and b, $d$, $n$ are taken as 10 and 5 respectively. From the analysis of hyperparameters, the best hyperparameters (n,d) for modeling the $R_a$ was found to be (5,10). The hyperparameter optimization is shown in Fig. 16. For GB and AB, we also have considered the same hyperparameters.  

\subsection{Gradient Boosting (GB)}
 Similar to random forests the basic building block of gradient boosted trees are the decision trees \citep{panda2022machine}.  However while random forests create random samples for training based on Bootstrap aggregation (or Bagging), other ensemble learning approaches utilize Boosting. In Boosting-based approaches, several weak learners are trained sequentially. The training of each subsequent tree is affected by the errors of the prior trees, where a higher weight is placed on samples where prior trees have been inaccurate. The boosting process is slower in comparison to the bagging method since in boosting the trees are added sequentially. The trees in boosting focus on errors from the previous one, which makes the algorithm self-correcting and accurate in the asymptotic limit.

\subsection{Adaptive Boosting (AB)}

The Adaboost regressor is a meta-estimator that starts by fitting a regressor on the actual dataset and then fits extra copies of the regressor on the same dataset but the weights of the instances are adjusted according to the error of the current prediction \citep{margineantu1997pruning}. The core principle of AdaBoost is to fit a series of weak learners on a repeatedly modified version of the data. The predictions of all of the weak learners are then combined through a weighted majority vote to produce a final prediction. 

\section{Germinal center algorithm for optimization of the machining process parameters}
The algorithm for the Germinal center optimization (GCO) technique is discussed here \citep{thieu_nguyen_2020_3711949,villasenor2018germinal}. It is a unique multivariate continuous optimization technique mainly inspired by the germinal center reaction. The germinal centers are formed by the accumulation of (B cells) lymphocytes B and other unsusceptible cells. These cells are bounded by some sluggish B cells and these sluggish B cells form when any type of infection persists. The antibody was developed to get the best harmony in germinal centers when B cells undergo a diversified and aggressive or ruthless process. The germinal center reaction is shown in figure 17. In germinal centers, there are two zones, i.e. dark and light. The procedures adopted for two zones are repeatedly influenced by the GCO technique which is mentioned in Algorithm 1 as described in the article. A flow chart of the GCO is presented in the figure 17. \ref{fig:gco_al} and discussed in following sub sections:

\begin{figure}
\centering
\includegraphics[width=1\textwidth]{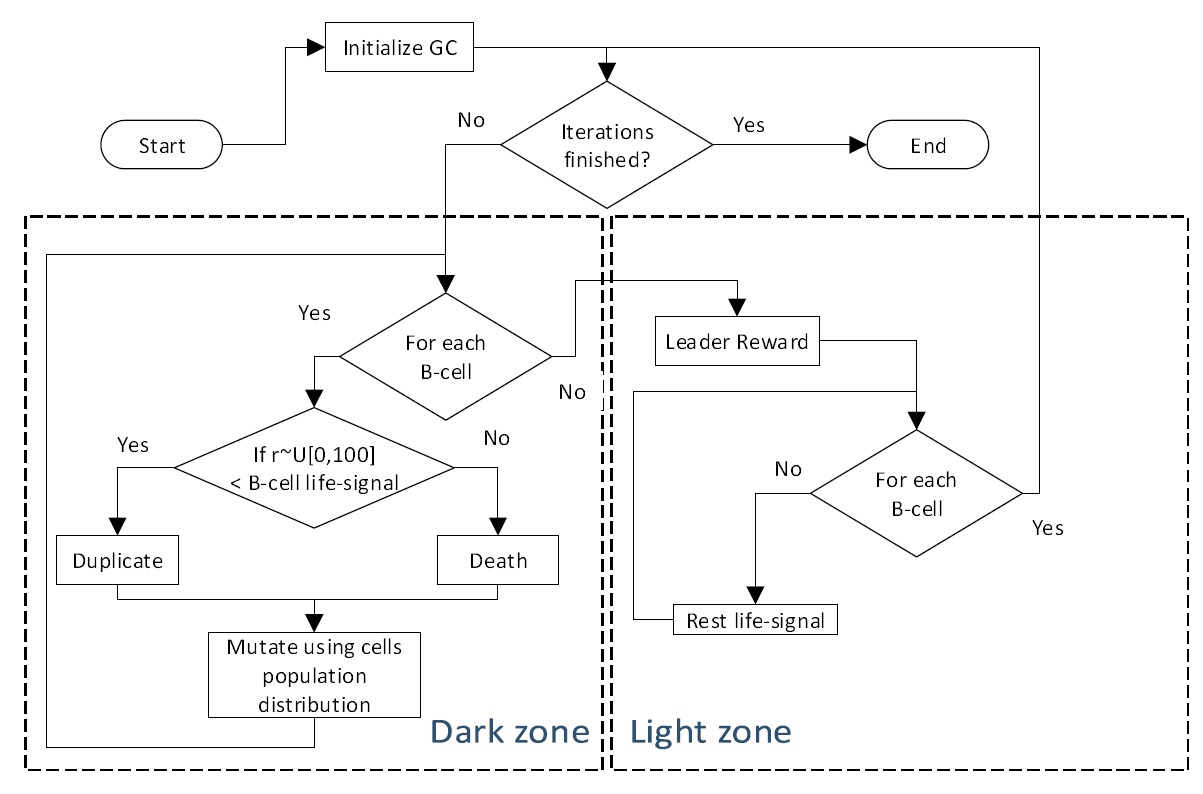}
\caption {The flow chart of germinal center algorithm reproduced from \cite{villasenor2018germinal}. \label{fig:gco_al}}
\end{figure}

\subsection{Process of dark zone}
The enhancement of B cells in the dark zone is mainly due to clonal expansion. And when new B cells formed through clonal expansion, some of them transformed through a process called hyper-mutation. The multiplicity of the B cells in our work is confirmed by the above two processes called clonal expansion and hyper-mutation. The solution for the multivariate optimization problem through germinal search is represented by B cells and Ab. 
Three providences a B shell can face in the dark zone like cell generation, cell death, and cell mutation.  Algorithms 2 and 3 are selected for this work. 

\subsection{Process of light zone}
The B cells are engaged in phagocytizing Ag in the light zone. The B cell incorporates the Ag and degrades in the proteins culture. When B cells are in the proteinous elements to a cell of T, the life signal was inhabited in the B cell. In our experimental analysis, the honor for the life signal L was the light zone. The amount of reward was decided based upon the performance of the solution. The requisite algorithm 4 was stated in the paper.
In general, the population distribution of the B cell is done through the competition in the germinal center optimization algorithm. The procedure adopted in this optimization technique is quite important. Because it’s different from DE, where uniform distribution is adopted. In uniform distribution, all the particles will get the chance for mutation. In the PSO technique, the mutation process is influenced by the leader’s behavior. In the GCO technique, the method was assumed by the population distribution. After that, the algorithm works like DE and PSO. When all the solutions will be having the same performance, the GCO technique is imprecise with DE. Similarly, when there will be one critical solution then the GCO technique is imprecise with PSO. By providing such methodologies, the exploration-exploitation can be balanced that changes its actions through the time span. The delayed leadership quality is also there. And this phenomenon occurred, when one normal particle will defeat the best particle. But its impact will be slowly reduced through no of iterations.
        
\section{Results and Discussion}
\begin{figure}
\centering
 \subfloat[]{\includegraphics[width=0.4\textwidth]{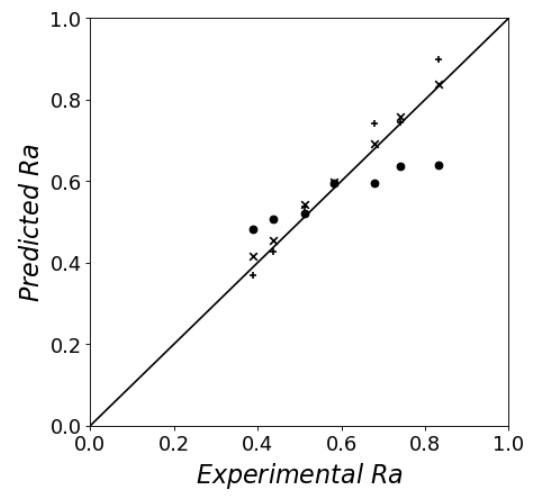}}
 \subfloat[]{\includegraphics[width=0.4\textwidth]{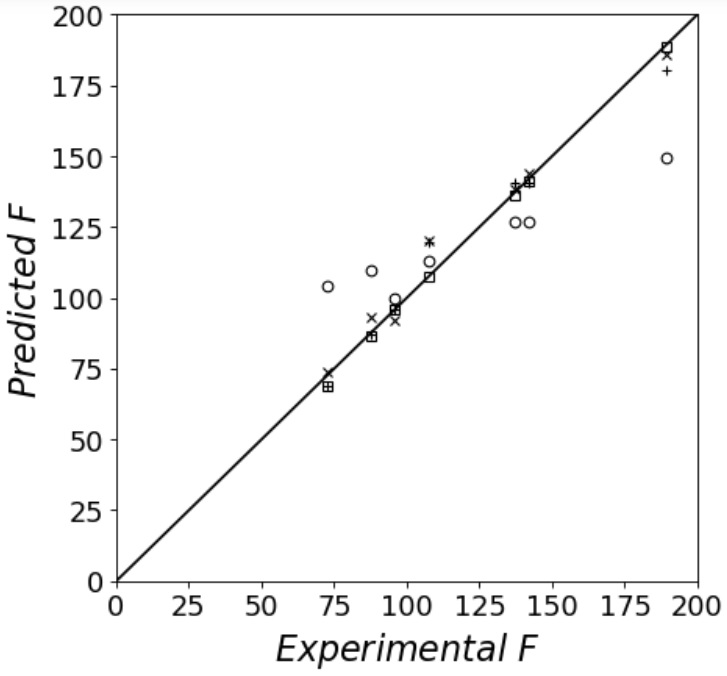}}\\
 \subfloat[]{\includegraphics[width=0.4\textwidth]{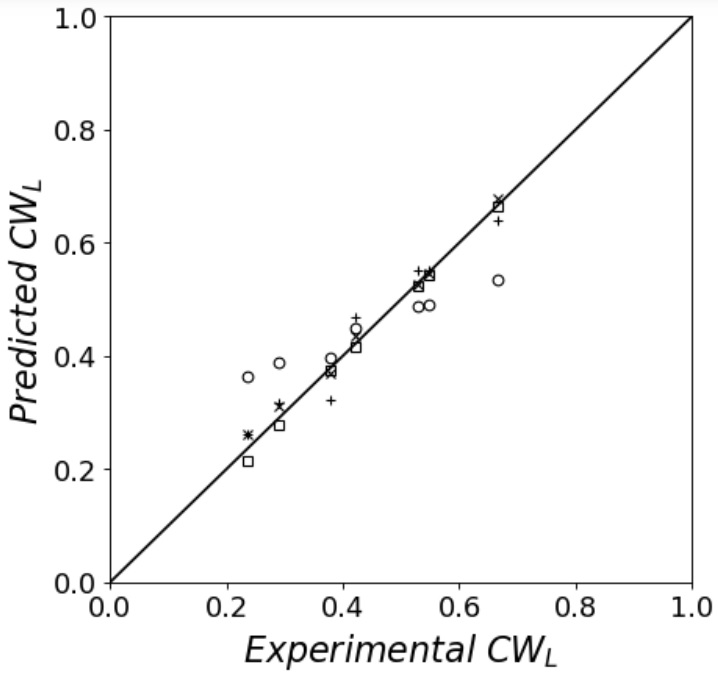}}
 \subfloat[]{\includegraphics[width=0.425\textwidth]{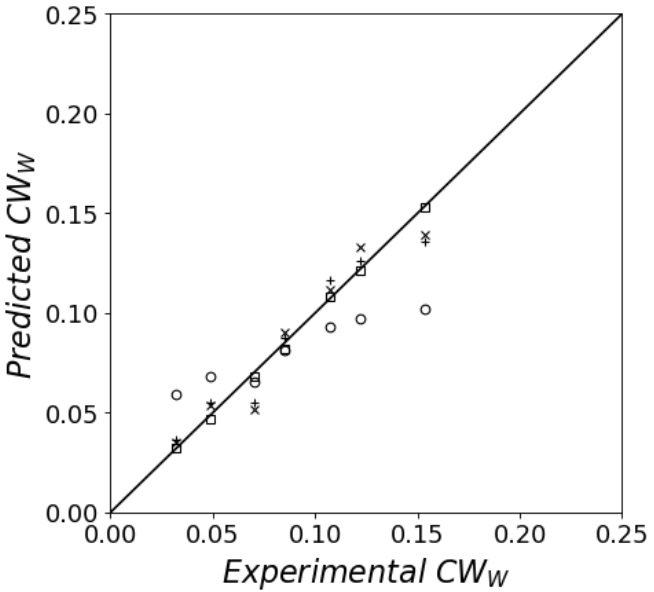}}\\
  \subfloat[]{\includegraphics[width=0.4\textwidth]{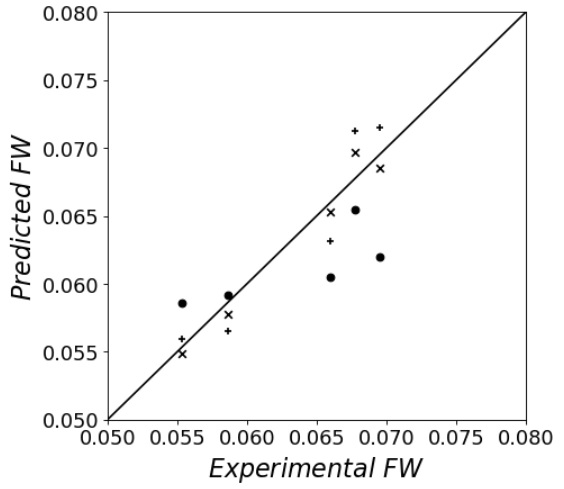}}
 
 \caption{Actual vs predicted values of the machining parameters for train-test dataset $D_1$. The predictions were made using ML models. square symbols PR, cross symbols RF, plus symbols AB and circles correspond to GB.\label{fig:ap_1}}
\end{figure}

\begin{figure}
\centering
 \subfloat[]{\includegraphics[width=0.4\textwidth]{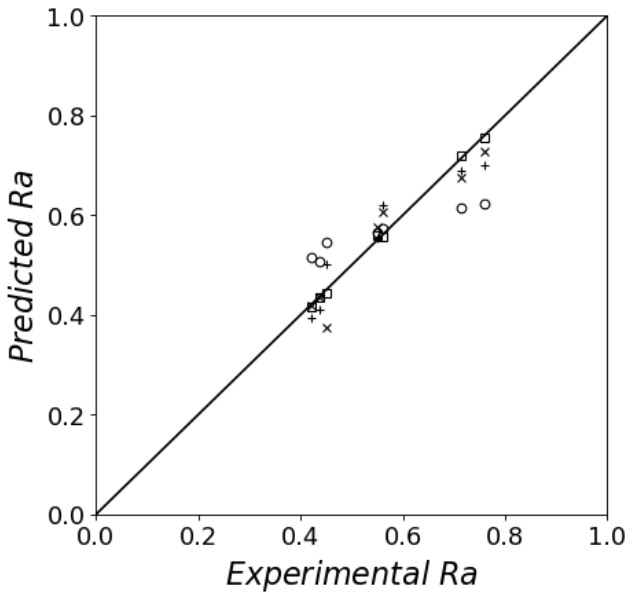}}
 \subfloat[]{\includegraphics[width=0.4\textwidth]{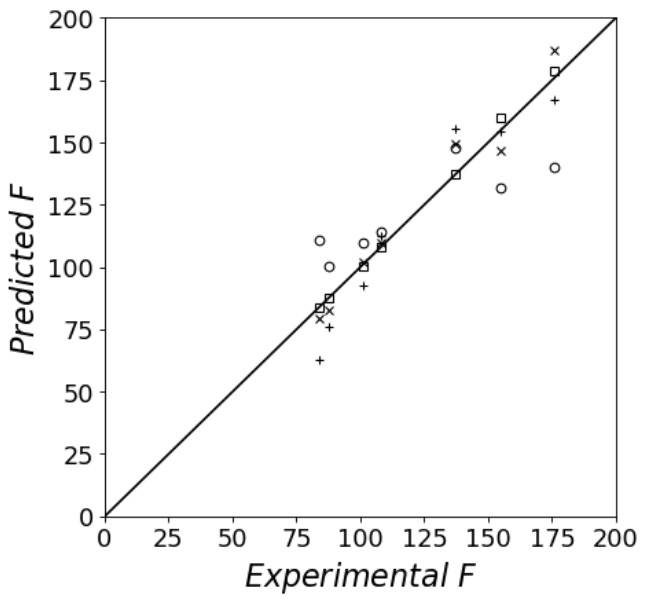}}\\
 \subfloat[]{\includegraphics[width=0.4\textwidth]{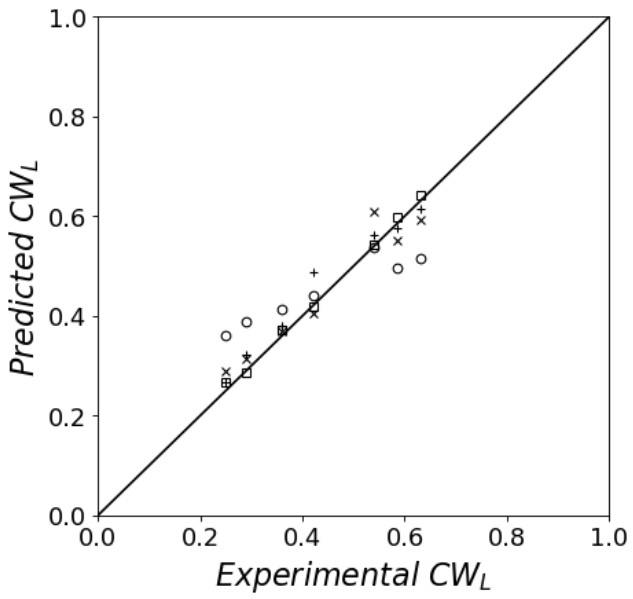}}
 \subfloat[]{\includegraphics[width=0.425\textwidth]{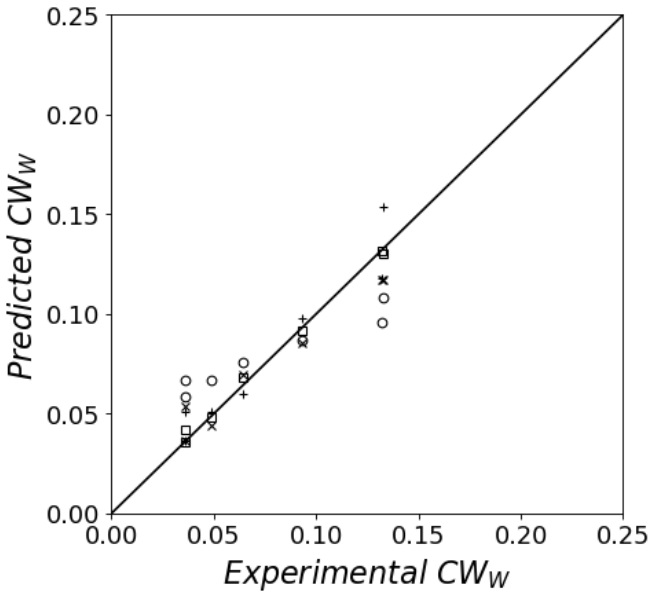}}\\
  \subfloat[]{\includegraphics[width=0.4\textwidth]{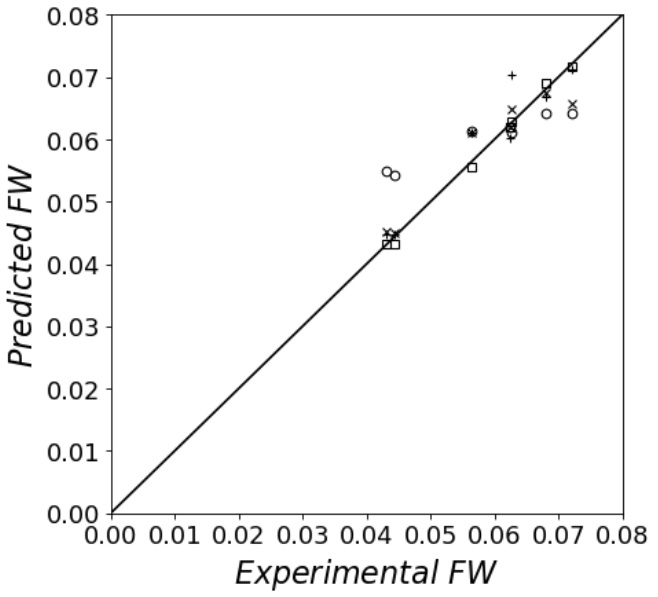}}
 
 \caption{Actual vs predicted values of the machining parameters for train-test dataset $D_2$. The predictions were made using ML models. square symbols PR, cross symbols RF, plus symbols AB and circles correspond to GB.\label{fig:ap_2}}
\end{figure}

In this section, the predictive capability of four ML models (polynomial regression, random forest, gradient boosting, and adaptive boosting) is discussed and compared based on their prediction accuracy. We have used three performance parameters for the development and comparison of the predictive capability of the ML models, those are $R^2$, $MSE$ and $MAE$ as defined in section \ref{MLmodels}. In table \ref{tbl:d1} and \ref{tbl:d2} the $R^2$, $MSE$ and $MAE$ values of different ML models are presented.Table \ref{tbl:d1} and \ref{tbl:d2} correspond to the train-test dataset $D_1$ and $D_2$ respectively. The $R^2$ values of all prediction of all the five parameters are found to be more than 0.95 for RF and PR for train-test dataset $D_1$. However, for the train-test dataset $D_2$ the $R^2$ values are greater than 0.95 only for PR. The $R^2$ values of results predicted through the other two ML algorithms e.g. ADA and GB is much lesser than 0.95, which signifies that those models are not suitable for predicting the machining parameters. Similar to $R^2$, $MSE$ and $MAE$ of predictions are small for PR. The actual vs predicted values of the predicted ML variables are presented in figures \ref{fig:ap_1} and \ref{fig:ap_2}.            

\subsection{ML based prediction of responses:}
For modeling the responses of $Ra$, $F$, $CW_L$, $CW_W$, $FW$ in terms of speed (s), feed (f), and depth of cut (d), we have used the equations generated by polynomial regression (PR) using the dataset $D1$, since PR can provide interpretable equations, those can be used in the development of the complex objective function, that is required to be used in the optimization of the process parameters. The equations learned through PR, are written as follows:   


\begin{equation}
\centering
\begin{split}
Ra=0.51-0.05s+0.19f+0.03d+0.09fs+0.0005sd-\\ 0.005fd-0.004s^2+0.006f^2+0.016d^2,\\
F=106.28-0.39s+33.69f+25.38d+21.95fs+2.91sd-\\4.22fd-22s^2+14.9f^2+21.38d^2,\\
CL=0.43+0.007s+0.12f+0.11d+0.06fs+0.008sd-\\0.0003fd-0.06s^2+0.03f^2+0.02d^2,\\
CW=0.07-0.02s+0.04f+0.02d-0.004fs-0.002sd+\\0.004fd-0.002s^2+0.002f^2+0.006d^2,\\
FW=0.06+0.004s+0.008f+0.004d+0.006fs-0.001sd-\\0.002fd+0.004s^2-0.003f^2-0.006d^2\\
\end{split}
\end{equation}

\begin{table}[htbp]
  \centering
  \begin{tabular}{|c|c|c|c|c|c|}
    \hline
     \cline{3-5}
     \multicolumn{2}{|c|}{}  & $R^2$ & MSE & MAE \\\hline
  \multirow{5}{*}{\rotatebox{90}{$Ra$}}  & PR & \(0.9916\) &  \(0.0002\) &  \(0.0119\) \\\cline{2-5}    
    &RF & \(0.9826\)  &    \(0.0003\) &  \(0.0184\) \\\cline{2-5}
   & ADA & \(0.9397\)   &  \(0.0013\) &  \(0.0276\) \\\cline{2-5}
   & GB & \(0.5645\)    & \(0.0099\) &    \(0.0813\) \\ 
   \hline  
   
    \multirow{5}{*}{\rotatebox{90}{$F$}}  & PR & \(0.9979\) &  \(2.85\) &  \(1.17\) \\\cline{2-5}    
    &RF & \(0.9638\)  &    \(49.58\) &  \(5.84\) \\\cline{2-5}
   & ADA & \(0.9011\)   &  \(135.69\) &  \(11.08\) \\\cline{2-5}
   & GB & \(0.6676\)    & \(455.98\) &    \(17.95\) \\ 
   \hline  
   
    \multirow{5}{*}{\rotatebox{90}{$CW_L$}}  & PR & \(0.9949\) &  \(0.0001\) &  \(0.0077\) \\\cline{2-5}    
    &RF & \(0.9886\)  &    \( 0.0002 \) &  \(0.0123\) \\\cline{2-5}
   & ADA & \(0.9613\)   &  \(0.0007\) &  \( 0.0264\) \\\cline{2-5}
   & GB & \(0.6531\)    & \(0.0068\) &    \(0.0721\) \\ 
   \hline  
   
   \multirow{5}{*}{\rotatebox{90}{$CW_W$}}  & PR & \(0.9981\) &  \(2.9E-06\) &  \(0.0014\) \\\cline{2-5}    
    &RF & \(0.9769\)  &    \(0.00003\) &  \(0.0053\) \\\cline{2-5}
   & ADA & \(0.9342\)   &  \(0.0001\) &  \(0.0083\) \\\cline{2-5}
   & GB & \(0.6381\)    & \(0.0005\) &    \(0.0194\) \\ 
   \hline  
   
   \multirow{5}{*}{\rotatebox{90}{$FW$}}  & PR & \( 0.9885 \) &  \(1.2E-06\) &  \(0.0008\) \\\cline{2-5}    
    &RF & \( 0.9865\)  &    \(1.4E-6\) &  \(0.0011\) \\\cline{2-5}
   & ADA & \(0.9519\)   &  \(5.1E-6\) &  \(0.0021\) \\\cline{2-5}
   & GB & \(0.5274\)    & \(4.9E-6\) &    \(0.0058\) \\ 
   \hline  
  \end{tabular}
  \caption{ML performance parameters for the train-test dataset $D_1$}
  \label{tbl:d1}
\end{table}
\begin{table}[htbp]
  \centering
  \begin{tabular}{|c|c|c|c|c|c|}
    \hline
     \cline{3-5}
     \multicolumn{2}{|c|}{}  & $R^2$ & MSE & MAE \\\hline
  \multirow{5}{*}{\rotatebox{90}{$Ra$}}  & PR & \(0.9974\) &  \(000000\) &  \( 0.0055\) \\\cline{2-5}    
    &RF & \(0.9002\)  &    \(0.0015\) &  \(0.0314\) \\\cline{2-5}
   & ADA & \(0.8933\)   &  \( 0.0016\) &  \(0.0361\) \\\cline{2-5}
   & GB & \(0.5260\)    & \(0.0074\) &    \(0.0751\) \\ 
   \hline  
   
    \multirow{5}{*}{\rotatebox{90}{$F$}}  & PR & \(0.9956\) &  \(4.6432\) &  \(1.3173\) \\\cline{2-5}    
    &RF & \(0.9457\)  &    \(57.7313\) &  \(6.3032\) \\\cline{2-5}
   & ADA & \( 0.8521\)   &  \(157.4010\) &  \(10.4949\) \\\cline{2-5}
   & GB & \(0.6094\)    & \(415.6819\) &    \(17.7077\) \\ 
   \hline  
   
    \multirow{5}{*}{\rotatebox{90}{$CW_L$}}  & PR & \(0.9947\) &  \(9.94e-05\) &  \(0.0086\) \\\cline{2-5}    
    &RF & \(0.9258\)  &    \(0.0014 \) &  \(0.0330\) \\\cline{2-5}
   & ADA & \( 0.9474\)   &  \(  0.0010\) &  \(  0.0266\) \\\cline{2-5}
   & GB & \(0.6467\)    & \( 0.0067\) &    \(0.0704\) \\ 
   \hline  
   
   \multirow{5}{*}{\rotatebox{90}{$CW_W$}}  & PR & \(0.9948\) &  \(7.78e-06\) &  \(0.002\) \\\cline{2-5}    
    &RF & \(0.9769\)  &    \(0.00003\) &  \(0.0053\) \\\cline{2-5}
   & ADA & \(0.9342\)   &  \(0.0001\) &  \(0.0083\) \\\cline{2-5}
   & GB & \(0.6381\)    & \(0.0005\) &    \(0.0213\) \\ 
   \hline  
   
   \multirow{5}{*}{\rotatebox{90}{$FW$}}  & PR & \( 0.9950 \) &  \(5.27e-07\) &  \(0.0005\) \\\cline{2-5}    
    &RF & \( 0.9033\)  &    \(1.02e-05\) &  \(0.0023\) \\\cline{2-5}
   & ADA & \(0.8737\)   &  \(1.34e-05\) &  \(0.0027\) \\\cline{2-5}
   & GB & \(0.5495\)    & \(4.79e-05\) &    \(0.0057\) \\ 
   \hline  
  \end{tabular}
  \caption{ML performance parameters for the train-test dataset $D_2$}
  \label{tbl:d2}
\end{table}

\subsection{Multi-objective optimization:}

Although we have developed the predictive machine learning models using different ensemble methods, those are characterized by complex non-linear functions, so it is not possible to use traditional optimization methods in conjunction with the ML methods. We have developed input-output correlation by using polynomial regression and the correlation was used to establish the objective function to be used in the multi-objective optimization. The objective the multi-objective optimization here is to minimize the $F$, $Ra$, $CW_L$, $CW_W$ and $FW$. The following objective function as shown in equation 7 has been proposed for the above-mentioned purpose:
\begin{equation}
COF(s,f,d)=w_1\frac{Ra}{Ra_{m}}+w_2\frac{F}{F_{m}}+w_3\frac{CW_L}{CW_{Lm}}+w_4\frac{CW_W}{CW_{Wm}}+w_5\frac{FW}{FW_{m}}
\end{equation}
where $w_{1}, w_{2}, w_{3}, w_{4}$ and $w_{5}$ weight values for the output parameters. Here, we have taken all the weight values as same, i.e. $w_{1}=w_{2}=w_{3}= w_{4}=w_{5}=1/5$. The minimization of the above complex objective function was performed based on constraints of machining operation. Three operation constraints were used to develop the complex objective function. The three constraints are the lower and upper limits of the experimental parameters. The constraints are as follows: 40 $\leq$ s $\leq$ 90, 0.04 $\leq$ f $\leq$ 0.16 and 0.2 $\leq$ d $\leq$ 0.5. There are different meta-heuristic methods such as genetic algorithm, differential evolution, particle swarm optimization, and simulated annealing available by which the above-mentioned problem can be solved. In this work, to solve the optimization problem a recently developed algorithm, named Germinal Center Optimization was used. The detailed theory of the Germinal center algorithm is available in \citep{villasenor2018germinal}. The optimal combination of input parameters (speed, feed, and depth of cut) for minimizing the outputs are found to be 60, 0.04, 0.2. 
\section{Concluding remarks}

After analyzing the experimental results, following conclusions are obtained. a) A columnar free dense structure coating was observed using scalable pulsed power plasma process, b) With 40, 50, 55 and 60m/min speed levels, cutting force was found to be diminished, whereas, when machining was conducted through 70, 80 and 90m/min sudden increment in cutting force was noticed. Mainly two wear mechanisms are observed i.e., abrasion and adhesion. Feed influenced the tool-chip contact length significantly. Both chip sticking and sliding was noticed on the tool rake face. Speed was observed to be the critical parameter for flank wear. Both morphological aspect and surface roughness was highly influenced by depth of cut and tool wear. Self-induced vibration observed at higher depth of cut. 

The experimental dataset was used for modeling the various machining parameters such as surface roughness, cutting force, flank wear, crater wear length, and crater wear width were modeled using machine learning algorithms (polynomial regression, random forest, gradient boost, and ADA boost). The predictive capability of four ML models was assessed in the modeling of the machining parameters. The performance parameters assessed are $R^2$, $MSE$, and $MAE$ of the predictions of the machining parameters. It was noticed that for both random forest and polynomial regression the $R^2$ value is greater than $0.9$, which signifies the fact that the learned correlation function maps the input and output parameters more accurately. However, ADA boost and gradient boost predictions are poor in comparison to RF and PR. In contrast to RF, PR predictions are comparatively better and the surrogate model learned through PR can be easily extracted and implemented in the formulation of the complex objective function. We recommend the use of PR for predictive modeling of the machining parameters. 

Hence we have used the PR learned surrogate models of different machining parameters in the development of the complex objective function that is required to be used in the germinal center-based optimization of the input parameters for the minimization of the output parameters. The optimal machining process parameters are found to be 60, 0.04, and 0.2 (speed, feed, and depth of cut respectively) that minimize the output parameters.

In the future course of work, in place of speed, feed and depth of cut,   tools of various shapes can be used for ML-based modeling and optimization in a machining process. The input features for various tools will be processed through shape parametrization using convolutional neural networks.
 
\bibliographystyle{apacite}
\bibliography{asmebib}


\clearpage

\end{document}